\documentclass[runningheads]{llncs}
\usepackage{graphicx}

\usepackage{tikz}
\usepackage{comment}
\usepackage{amsmath,amssymb} 
\usepackage{color}
\usepackage{floatrow}
\usepackage{booktabs}
\usepackage{multirow, multicol}
\usepackage{float}
\usepackage{subcaption}
\usepackage{hyperref} 
\usepackage{pifont}
\newcommand{\cmark}{\ding{51}}%
\newcommand{\xmark}{\ding{55}}%
\usepackage{orcidlink}

\DeclareMathOperator*{\argmin}{arg\,min}
\newcommand{\placeholder}{NSA}

\usepackage[capitalize]{cleveref}
\crefname{section}{Sec.}{Secs.}
\Crefname{section}{Section}{Sections}
\Crefname{table}{Table}{Tables}
\crefname{table}{Tab.}{Tabs.}

\usepackage[accsupp]{axessibility}  

\begin{document}
\pagestyle{headings}
\mainmatter
\def\ECCVSubNumber{7519}  

\title{Natural Synthetic Anomalies for Self-Supervised Anomaly Detection and Localization} 

\titlerunning{Natural Synthetic Anomalies}
%
\author{Hannah M. Schl\"{u}ter\inst{1}\orcidlink{0000-0002-5965-7889}\index{Schl\"{u}ter, Hannah M.} \and
Jeremy Tan\inst{1}\orcidlink{0000-0002-9769-068X} \and
Benjamin Hou\inst{1}\orcidlink{0000-0003-3968-1707} \and \\
Bernhard Kainz\inst{1,2}\orcidlink{0000-0002-7813-5023}}
\authorrunning{H. Schl\"{u}ter et al.}
%
\institute{Imperial College London
\email{\{hannah.schlueter17,j.tan17,benjamin.hou11,b.kainz\}@imperial.ac.uk} \and
Friedrich-Alexander-Universit\"at Erlangen-N\"urnberg}
\maketitle

\begin{abstract}
    We introduce a simple and intuitive self-supervision task, Natural Synthetic Anomalies (\placeholder{}), for training an end-to-end model for anomaly detection and localization using only normal training data. \placeholder{} integrates Poisson image editing to seamlessly blend scaled patches of various sizes from separate images. This creates a wide range of synthetic anomalies which are more similar to natural sub-image irregularities than previous data-augmentation strategies for self-supervised anomaly detection. We evaluate the proposed method using natural and medical images. Our experiments with the MVTec AD dataset show that a model trained to localize \placeholder{} anomalies generalizes well to detecting real-world a priori unknown types of manufacturing defects. Our method achieves an overall detection AUROC of \textbf{97.2} outperforming all previous methods that learn without the use of additional datasets. Code available at \url{https://github.com/hmsch/natural-synthetic-anomalies}.\medskip
    
\end{abstract}

\section{Introduction}
Anomaly detection is a binary classification task where the aim is to separate normal data from anomalous examples. There are different types of anomaly detection depending on what training data and labels are available. A difficult yet realistic setting is to detect and localize unknown types of anomalies while only having access to normal data during training. 
To be useful in real applications, an automated system must be able to detect 
subtle and rare anomalies; irregularities that are either impossible to spot for humans because of contextual uniformity or get lost due to task-remote stimuli that lead to inattentional blindness~\cite{mack1998inattentional}.

Attempting to detect rare anomalies often means that it is impossible to acquire sufficient amounts of human-annotated training data for a supervised method. Obtaining precise ground-truth annotations is time-consuming and requires expert knowledge depending on the application domain. Anomaly detection based on only normal data has applications in many areas including defect detection in industrial production pipelines \cite{BergmannPaul2021TMAD,cohen2021subimage,defard2020padim,li2021cutpaste,rippel2020modeling,app10238660}, unsupervised lesion detection in medical images \cite{marimont2020anomaly,Pawlowski2018UnsupervisedLD,SCHLEGL201930,tan2020detecting,PII_miccai,zimm_VAE}, or finding unusual events in surveillance videos \cite{DBLP:conf/cvpr/AbatiPCC19,cohen2021subimage,defard2020padim}. \par 

The main challenge of unsupervised approaches is designing a training setup that will encourage the model to learn features relevant to anomaly detection without having any prior knowledge of the types of anomalies to expect. A common theme among top-performing approaches \cite{defard2020padim,gudovskiy2021cflowad,rippel2020modeling,Roth_2022_CVPR} for anomaly detection in natural images, specifically on the MVTec AD benchmark, is to sidestep this challenge by relying on deep features from pre-trained ImageNet models. 

We find approaches that learn from scratch more interesting. They are more widely applicable to other domains where the usefulness of ImageNet pre-training is limited, such as medical imaging~\cite{NEURIPS2019_eb1e7832}. Many 
rely on learning a compressed representation of normal data and use these embeddings or reconstructions derived from them 
to define an anomaly score. Self-supervised learning is thus becoming a prominent strategy in anomaly detection. By designing an appropriate task, self-supervision can be an effective proxy for supervised learning, bypassing the need for labeled data. While various self-supervised tasks, such as context prediction \cite{7410524} or estimating geometric transformations \cite{gidaris2018unsupervised,10.5555/3327546.3327644}, can be used to learn a compressed representation of the data, recent works \cite{li2021cutpaste,tan2020detecting,PII_miccai,zavrtanik2021draem} show that data-augmentation strategies mimicking real defects are particularly effective for sub-image anomaly detection. These methods create synthetic anomalies by replacing or blending image patches with content from other images or image locations. However, 
recently proposed synthesis strategies~\cite{li2021cutpaste,zavrtanik2021draem}
feature obvious discontinuities. This is raising concerns that the model may overfit to  prior assumptions that are inherently encoded in synthetic manipulations. To prevent this, one can either resort to simpler encoders \cite{li2021cutpaste}, which impedes end-to-end localization, or add additional networks for reconstruction \cite{zavrtanik2021draem}, which greatly increases model size, computational costs, and training time. Even methods that linearly interpolate similar patches to create more subtle irregularities can suffer from the same problem~\cite{tan2020detecting}.  \cite{PII_miccai} solves this problem by using Poisson image editing \cite{10.1145/1201775.882269}, but the resulting anomalies are so subtle that they may represent variations of the normal class rather than true anomalies. 

\noindent\textbf{Contribution.} We introduce a simple and intuitive self-supervised method for sub-image anomaly detection and localization. Our Natural Synthetic Anomalies (\placeholder{)}, are a) more natural than the current state-of-the-a-art CutPaste~\cite{li2021cutpaste}, FPI~\cite{tan2020detecting}, or DRAEM~\cite{zavrtanik2021draem} due to the use of Poisson image editing, b) more diverse than CutPaste, FPI, or PII anomalies due to rescaling, shifting and a new Gamma-distribution-based patch shape sampling strategy, and c) more relevant to the task by imposing background constraints and using pixel-level labels derived from the resulting difference to the normal image rather than interpolation factors as used in FPI and PII. 
Like FPI and PII, \placeholder{} can be used to train an end-to-end model for anomaly detection and localization rather than generating compressed representations for a multi-stage pipeline. \par 

We evaluate the proposed method on the MVTec AD dataset \cite{BergmannPaul2021TMAD} which contains normal training data and both normal and anomalous test data for a wide range of natural and manufacturing defects for 10 object and 5 texture classes. \placeholder{} achieves the new state-of-the-art localization (96.3 AUROC) and dectection (97.2 AUROC) performance among methods that do not use additional datasets. It also performs comparably to the best methods that use additional data and much bigger models  \cite{defard2020padim,zavrtanik2021draem}. \cite{defard2020padim} uses a model pre-trained on ImageNet. Compared to the large amount of data in ImageNet, our method only uses MVTec AD data, which 
contains between 60 and 391 training images per class. \par 

\placeholder{} is a very general method for creating diverse and realistic synthetic anomalies in images. Since it does not rely on pre-training with ImageNet or any other dataset, it can easily be adapted to domains beyond natural images. Thus, we also evaluate \placeholder{} using a curated subset of a public chest X-ray dataset \cite{wang2017chestx} where it outperforms other state-of-the-art self-supervised methods for disease detection. 

\section{Related Work}

\noindent\textbf{Reconstruction-based anomaly detection} establishes pixel-level and image-level anomaly scores from the pixel-wise reconstruction error using variational autoencoders (VAE) \cite{zimm_VAE}, Bayesian autoencoders \cite{Pawlowski2018UnsupervisedLD}, generative adversarial networks (GAN) \cite{SCHLEGL201930}, or the restoration distance using a vector-quantized VAE (VQ-VAE) \cite{marimont2020anomaly,app10238660} trained with normal data. Anomaly scores can be improved by leveraging additional information derived from the model, such as the discriminator output when using a GAN \cite{SCHLEGL201930}, the KL-divergence of the latent representation of a VAE for image-level scores or its gradient for pixel-level scores \cite{zimm_VAE}, or the likelihood of the latent representation under a learnt prior using a VQ-VAE \cite{marimont2020anomaly,app10238660} or latent space autoregression \cite{DBLP:conf/cvpr/AbatiPCC19}. A downside of these approaches is that it is difficult to control the capacity of the model. Depending on regularization, the model cannot reconstruct all details of normal examples well or may be able to reconstruct anomalous regions too.

\noindent\textbf{Embedding-based anomaly detection} derives an anomaly score from the distance between embedding vectors of normal training images and test examples. An embedding-similarity metric can be defined using any method for one-class classification such as support vector data description (SVDD) used in \cite{Yi_2020_ACCV}, Gaussian distributions used in \cite{defard2020padim,li2021cutpaste,rippel2020modeling}, or nearest neighbour search used in \cite{cohen2021subimage,Roth_2022_CVPR}. Features for the embedding vectors are often extracted from pre-trained deep neural networks \cite{cohen2021subimage,defard2020padim,rippel2020modeling,Roth_2022_CVPR}, but can also be learned from scratch using a self-supervised task \cite{li2021cutpaste} or together with the one-class classification objective in Deep-SVDD \cite{pmlr-v80-ruff18a} and Deep One-Class Classification (DOCC) \cite{ruff2021unifying} or a combination thereof \cite{Yi_2020_ACCV}.
When using embeddings of the entire image, embedding-based approaches can perform detection but not localization and are hence less interpretable. To circumvent this issue, \cite{defard2020padim,li2021cutpaste,Roth_2022_CVPR,Yi_2020_ACCV} work with patch-level embeddings to create an anomaly map while \cite{cohen2021subimage} compares test images to their nearest neighbors from the training set at the pixel-level.
In a similar vein, flow-based methods can perform density estimation~\cite{DBLP:conf/iclr/GrathwohlCBSD19}. However, these methods can sometimes assign higher likelihood to outlier samples~\cite{DBLP:conf/iclr/NalisnickMTGL19} and typically need pre-trained feature extractors to achieve better performance~\cite{DBLP:journals/corr/abs-2111-07677Fastflow}. 

\noindent\textbf{Self-supervised learning.} A supervisory signal from a proxy task defined based on unlabeled data, such as predicting the relative position of patches \cite{7410524} or estimating geometric transformations \cite{gidaris2018unsupervised,10.5555/3327546.3327644}, can help the model learn useful features for a downstream task. While \cite{7410524,gidaris2018unsupervised,10.5555/3327546.3327644} use features learned from the proxy task to discover different object classes, self-supervised learning has also been successfully applied to sub-image anomaly detection \cite{li2021cutpaste,tan2020detecting,PII_miccai,Yi_2020_ACCV,zavrtanik2021draem}. In \cite{tan2020detecting,PII_miccai}, the output for the self-supervised task is a prediction of the interpolation factor where a foreign patch from another observation in the training distribution has been blended into the current image. This output is used directly as an anomaly score without any further training step. We also employ this general setup for our method. In \cite{zavrtanik2021draem}, the foreign patches are first removed by a reconstructive sub-network and then a discriminative sub-network produces an anomaly map by comparing the input image with foreign patches to the reconstruction.


\noindent\textbf{Poisson image editing.} Pasting part of one image into another causes obvious discontinuities. \cite{10.1145/1201775.882269} developed a method to seamlessly clone an object from one image into another image.  
For a source image given by $g$ and a destination image given by $f^*$, we seek an interpolant $f$ over the interior of a region $\Omega$ with boundary $\partial \Omega$ that solves the minimization problem given by \eqref{eq:pie_min}. According to \cite{10.1145/1201775.882269}, this has the unique solution of the Poisson partial differential equations \eqref{eq:pie_poisson} with Dirichlet boundary conditions given by the destination image.
\begin{align}
    f &= \argmin_f \iint_\Omega \vert \nabla f - \mathbf{v} \vert^2 \; \text{with} \; f\vert_{\partial \Omega} = f^*\vert_{\partial \Omega}
    \label{eq:pie_min} \\
    \Delta f &= \text{div} \mathbf{v} \; \text{over} \; \Omega, \;\text{with} \; f\vert_{\partial \Omega} = f^*\vert_{\partial \Omega}
    \label{eq:pie_poisson}
\end{align}
\cite{10.1145/1201775.882269} gives two options for defining the guidance field $\mathbf{v}$: a) use the source image gradient \eqref{eq:grad_src} or b) a mix of source and destination gradients \eqref{eq:grad_mix}.
\begin{align}
    \mathbf{v} &= \nabla g
    \label{eq:grad_src} \\
    \forall \mathbf{x} \in \Omega, \mathbf{v}(\mathbf{x}) &= \begin{cases} \nabla f^*(\mathbf{x}), &\text{if}\; \vert\nabla f^*(\mathbf{x})\vert > \vert\nabla g(\mathbf{x})\vert, \\ \nabla g(\mathbf{x}), &\text{otherwise}\end{cases}
    \label{eq:grad_mix}
\end{align}
In practice, a finite difference discretization of \eqref{eq:pie_poisson} is solved numerically. Seamless cloning is implemented in the OpenCV library \cite{opencv_library} which we use in our self-supervised task. A Poisson image editing approach has also recently been used for anomaly detection in medical images~\cite{PII_miccai}.

\section{\placeholder{} Self-supervised task} \label{sec:nsa}
As only normal data is available at training time, the model needs to be trained using a proxy task.
In our case, the task is to localize synthetic anomalies created from normal data by blending a patch from a source image into the destination image as follows: 
\begin{enumerate}
\setlength\itemsep{0em}
    \item Select a random rectangular patch in the source image.
    \item Randomly resize the patch and select a different destination location.
    \item Seamlessly blend the patch into the destination image.
    \item Optionally, repeat steps 1-3 to add multiple patches to the same image.
    \item Create a pixel-wise label mask.
\end{enumerate}

\begin{figure}
\centering
  \includegraphics[width=0.7\textwidth]{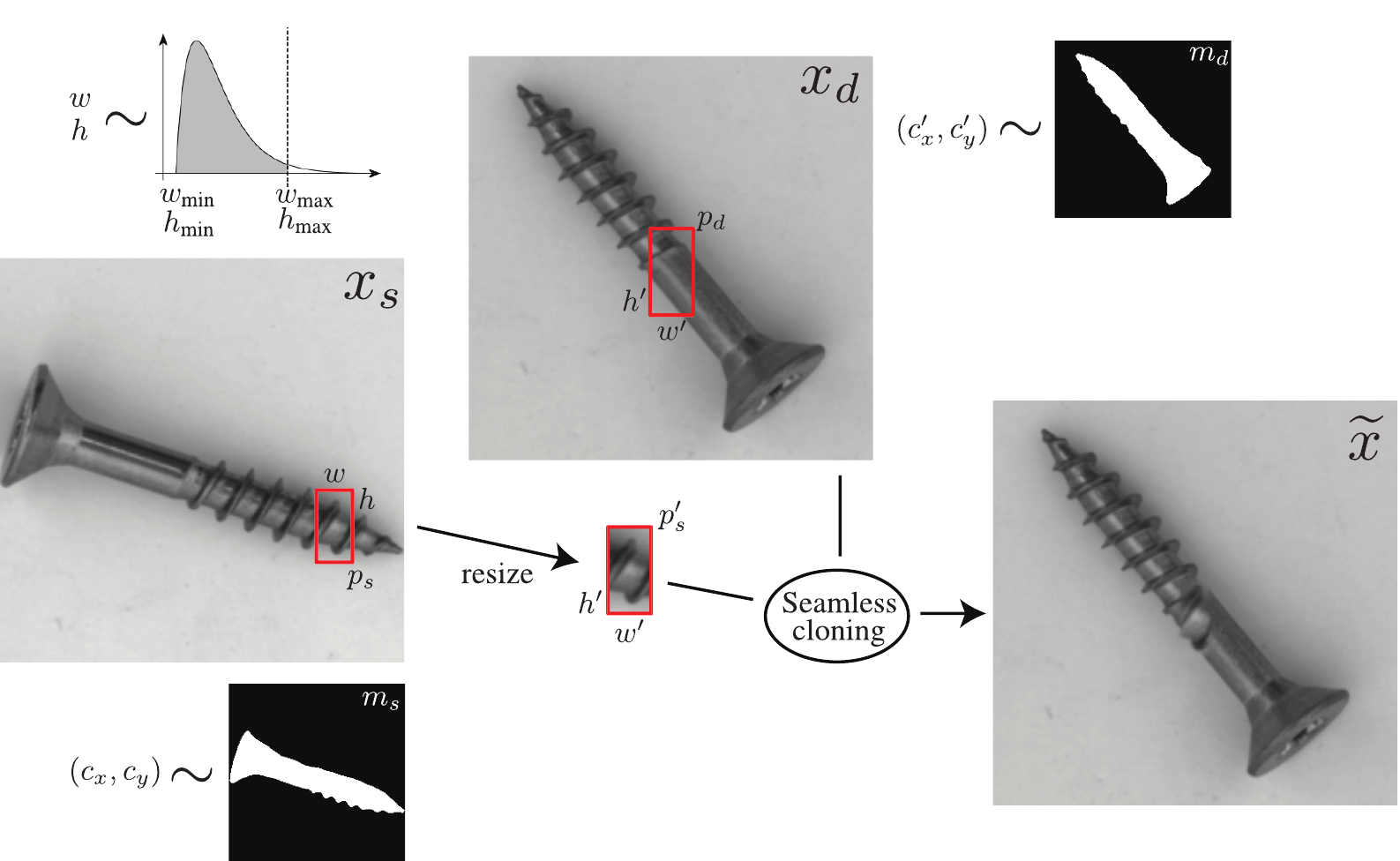}
  \caption{\placeholder{} anomalies are created by seamlessly cloning a patch from a normal training image into another normal training image.}
  \label{fig:diagram}
\end{figure}

\noindent\textit{Patch sampling and constraints:} Given two normal $W\times H$ training images $x_s$ and $x_d$, we select a random rectangular patch $p_s$ with width $w$ and height $h$ sampled from a truncated Gamma distribution, and center $(c_x, c_y)$ in the source image $x_s$ sampled from a uniform distribution:
\begin{align}
 w &= W\min\left(\max\left(w_{\text{min}},  0.06 + r_w\right), w_{\text{max}}\right), \quad \text{with} \quad r_w \sim \text{Gamma}(2, 0.1)  \\
    h &= H\min\left(\max\left(h_{\text{min}}, 0.06 + r_h\right), h_{\text{max}}\right), \quad \text{with} \quad r_h \sim \text{Gamma}(2, 0.1) \\
    c_x &\sim U\left(W \frac{w_{\text{min}}}{2}, W - W \frac{w_{\text{min}}}{2}\right), \quad
    c_y \sim U\left(H \frac{h_{\text{min}}}{2}, H - H \frac{h_{\text{min}}}{2}\right)\label{eq:center_src}
\end{align}
Sampling the width and height from a truncated Gamma distribution means we assume anomalies are local (small) but want the model to be able to recognize larger irregularities too. Hence, some long slim rectangles and occasionally large patches are generated. The width and height bounds are selected based on the dimensions of the object. For images containing an object and a plain background, we calculate object masks $m_s$ and $m_d$ by thresholding the pixel-wise absolute difference to the background brightness $b$. For each pixel $i$ the masks are given by: 
\begin{align}
     m_s^{(i)} = | x_s^{(i)} - b | < t_\text{brightness}, \quad  m_d^{(i)} = | x_d^{(i)} - b | < t_\text{brightness} 
\end{align}
We apply \eqref{eq:center_src} repeatedly until $(p_s \cap m_s) / (wh) > t_\text{object}$ to ensure the patch contains part of the object. 
Then we resize the patch to obtain $p'_s$ with width $w'=sw$ and height $h'=sh$. We select a destination patch $p_d$ in the destination image $x_d$ with the same dimensions and center $(c'_x, c'_y)$ where:
\begin{align}
    s &= \max\left(\frac{w_\text{min}}{w}, \frac{h_\text{min}}{h}, \min\left(r_s, \frac{w_\text{max}}{w}, \frac{h_\text{max}}{h}\right)\right) \quad \text{with} \quad r_s \sim N(1, 1/4)     \\ 
    c'_x &\sim U\left(W \frac{w'}{2}, W - W \frac{w'}{2}\right), \quad c'_y \sim U\left(H \frac{h'}{2}, H - H \frac{h'}{2}\right)
    \label{eq:center_dest}
\end{align}
To prevent creating many examples of patches floating in the background, we apply \eqref{eq:center_dest} repeatedly until
\begin{align}
(p_d \cap m_d) / (w'h') &> t_\text{object} \quad &\text{(contains part of the object),} \\ 
(m_{p_d} \cap m_{p'_s})/ |m_{p'_s}| &> t_\text{overlap} \quad &\parbox{5cm}{(object portions of patch and destination image overlap)}
\end{align}
where $m_{p_d}$ and $m_{p'_s}$ are the object masks of the source and destination patches. We seamlessly blend $p'_s$ into $x_d$ at location $(c'_x, c'_y)$ to obtain the training sample $\widetilde x$. After blending the first patch, we add up to $n-1$ further patches by flipping $n-1$ coins whether to add another patch or not. \cref{fig:diagram} shows a simplified outline of how the synthetic anomalies are created. \par

\noindent\textit{Hyperparameters as assumptions:} The patch selection procedure has a number of hyperparameters which characterize our synthetic out-distribution. Ideally, we would want this synthetic out-distribution to match the real out-distribution as closely as possible but since the real distribution is unknown at training time, we cannot use it to select the hyperparameters and sampling distributions. As \cite{pmlr-v139-zhang21g} proved for generative models, no test statistic is useful for all possible out-distributions and some assumptions must be made to increase the likelihood that the test is useful for relevant or probable out-distributions. For our method, we try to keep these assumptions as broad as possible by generating many different sizes, locations, quantities, and shapes of anomalies and only disregarding those which are obviously irrelevant due to extreme sizes compared to the object or little overlap with the object. \par

\noindent\textit{Labels:} We use the local intensity differences where a foreign patch has been introduced to create a pixel-wise label $\widetilde y$ which is either a) binary: whether there is a difference or not, b) continuous based on the mean absolute intensity difference across $C$ color channels, or c) a logistic function of the previous. All labels are median filtered to be more coherent. Before filtering, the label values at each pixel $i$ are calculated as follows:
\begin{align}
    \widetilde y^{(i)}_\text{binary} = \begin{cases}
    1, & \text{if } \widetilde x^{(i)} \neq x_d^{(i)}\\
    0,              & \text{otherwise}
    \end{cases} &, \quad\quad
    \widetilde y^{(i)}_\text{continuous} =  \frac{1}{C}\sum_{c=1}^C|\widetilde x^{(i,c)} - x_d^{(i,c)}| \\
    \widetilde y^{(i)}_\text{logistic} &= \dfrac{\widetilde y^{(i)}_\text{binary}}{1 + \exp\left(-k\left(\widetilde y^{(i)}_\text{continuous} - y_0\right)\right)}
\end{align}
In contrast, FPI \cite{tan2020detecting} and PII \cite{PII_miccai} use the patch interpolation factor as a label. This is somewhat ill-posed because the interpolation factor cannot be determined without knowing the pixel intensities of both the source and destination patches. Our labels are directly related to the change in intensity (created by the patch blending) and therefore provide a more consistent training signal. 
\par

\noindent\textit{Loss:} When using bounded labels ($\widetilde y_\text{binary}$ or $\widetilde y_\text{logistic}$) we define our pixel-wise regression objective using binary cross-entropy loss. For unbounded labels ($\widetilde y_\text{cont.}$) we use mean squared error loss. The loss is given in \eqref{eq:loss_bce}--\eqref{eq:loss_mse} where $\widehat y = f(\widetilde x)$ is the output of a deep convolutional encoder-decoder.
\begin{align}
\label{eq:loss_bce}
\mathcal{L}_{\text{bce}} = \frac{1}{W \times H}\sum_{i} &-\widetilde y^{(i)}_\text{bounded} \log \widehat y^{(i)}  - (1 - \widetilde y^{(i)}_\text{bounded}) \log \left(1 - \widehat y^{(i)} \right)  \\   
\mathcal{L}_{\text{mse}} = \dfrac{1}{W \times H}\sum_{i} &\left(\widetilde y^{(i)}_\text{continuous} -  \widehat y^{(i)}\right)^2
\label{eq:loss_mse}
\end{align}
\begin{figure}
  \centering
  \includegraphics[width=1.0\textwidth]{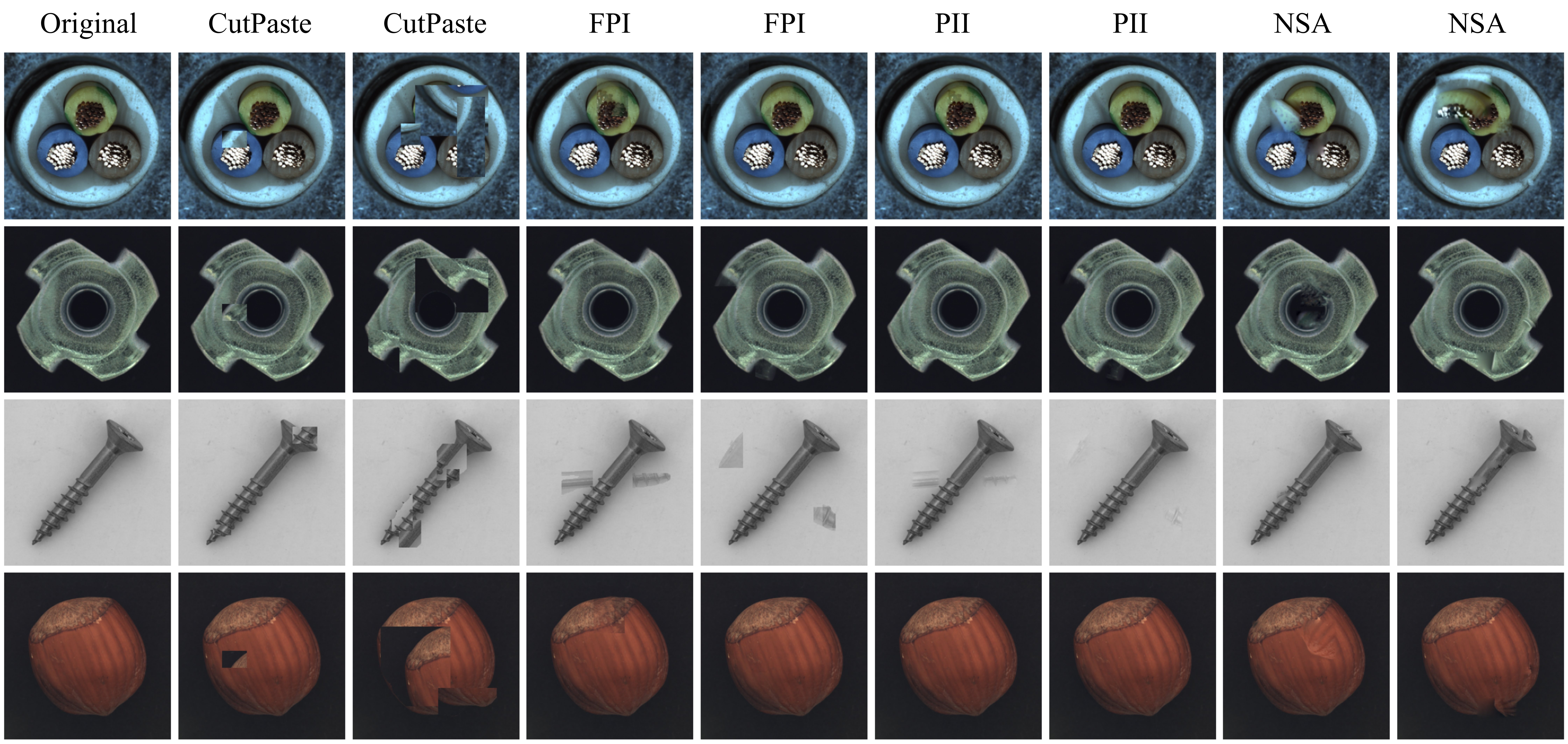}
  \caption{Synthetic anomalies created with CutPaste, FPI, PII, and \placeholder{}.}
  \label{fig:self_sup_examples}
\end{figure}
By varying size, aspect ratio, source and destination location, and resizing the scale of the patches, this method dynamically creates a wide range of synthetic anomalies during training. The examples feature changes in size, shape, texture, location, and color of local image components as well as missing components by blending in a patch containing some background, while staying true to the overall distribution of the images and avoiding obvious discontinuities. Hence, these examples are a more realistic approximation of natural sub-image anomalies than CutPaste augmentations constructed by simply pasting patches at different locations \cite{li2021cutpaste} and more diverse than interpolating patches from two separate images at corresponding locations as in FPI \cite{tan2020detecting} and PII \cite{PII_miccai} although still noticeably artificial to a human observer (\cref{fig:self_sup_examples}). \par 

\section{Experiments}
We compare end-to-end detection and localization models trained using our self-supervised task to end-to-end models trained using our implementations of FPI \cite{tan2020detecting}, PII \cite{PII_miccai}, and CutPaste augmentation \cite{li2021cutpaste} on the MVTec AD dataset \cite{BergmannPaul2021TMAD} and a curated subset of a public chest X-ray dataset \cite{wang2017chestx}. We assess performance using the area under the receiver operating characteristic curve (AUROC). \par

\noindent\textit{Datasets:} MVTec AD \cite{BergmannPaul2021TMAD} contains normal training data and normal and anomalous test data featuring various types of natural and manufacturing defects for 10 object and 5 texture classes.\par 
The NIH chest X-ray dataset \cite{wang2017chestx} contains normal images as well as 14 different types of pathological patterns. There is a lot of natural variation in the normal class which is challenging for unsupervised methods. However, the most obvious differences are easily explained by the different views and the gender of the patients. We reduce this variation by reducing the curated subset defined in \cite{tang2020automated} further to only posteroanterior (back-to-front) view images of patients aged over 18 and separating them by gender. This leaves us with 1973 normal training images, 299 normal and 139 abnormal test images of male patients. For female patients, we have 1641 normal training, 244 normal and 123 abnormal test images. We call this dataset re-curated chest X-ray (rCXR) in the following. Note that the authors of PII \cite{PII_miccai} used the full NIH chest X-ray rather than the curated subset defined in \cite{tang2020automated}. 

\subsection{Network architecture and training setup}

We use an encoder-decoder architecture with ResNet-18 \cite{7780459} without the classification layers as the encoder, two 1x1 convolutions in the bottleneck to reduce the number of channels and a simpler ResNet-based decoder. The final activation is sigmoid and we use binary-crossentropy loss for all models besides \placeholder{} (continuous) for which we use ReLU activation and mean squared error loss as the labels are unbounded. The models are trained on batches of size 64 using Adam \cite{DBLP:journals/corr/KingmaB14} with a cosine-annealing learning rate \cite{DBLP:conf/iclr/LoshchilovH17} that decays from $10^{-3}$ to $10^{-6}$ over 320 epochs. For non-aligned objects, the loss takes longer to converge, so we use 560 epochs for the hazelnut, metal nut, and screw classes in the MVTec AD dataset. For rCXR, we use 240 epochs. The same training hyperparameters are used for all variants of the self-supervised task. Hyperparameters for the self-supervised task are given in the supplementary material. Note that in our implementation of FPI, PII, and CutPaste we use object masks and the patch sizes are sampled from a truncated Gamma distribution rather than a uniform distribution \cite{li2021cutpaste,tan2020detecting,PII_miccai} to allow for a more fair comparison with \placeholder{}. We call our CutPaste baseline CutPaste (end-to-end) to distinguish it from the multi-stage framework used in \cite{li2021cutpaste}. \par 
The MVTec AD images have high resolutions of up to $1024\times 1024$ pixels and use the RGB color scheme. We resize object images to $256 \times 256$ pixels, apply a random rotation of up to 5 degrees for non-aligned and rotation invariant objects (bottle, hazelnut, metal nut, screw), center-crop to $230\times 230$ pixels and crop a random $224\times 224$ part of the image before creating self-supervised training examples to achieve slight rotation and translation invariance. When testing we use $224\times 224$ center-crops of $256\times 256$ object images. For texture classes, we use random $256\times 256$ crops of $264\times 264$ images for training and $256\times 256$ images for testing. Intensities are normalized using the mean and standard deviation of ImageNet as commonly used before feeding them into the model. \par
The rCXR images have a resolution of $1024\times 1024$ pixels in grayscale. We resize them to $256 \times 256$ pixels for training and apply a random rotation of up to 3 degrees, center-crop to $230\times 230$ pixels and take a random crop of $224\times 224$ pixels. For testing we use $224\times 224$ center-crops of $256 \times 256$ resampled images.

\noindent\textit{Implementation:}\label{sec:impl} We use PyTorch \cite{NEURIPS2019_9015} V1.8.1 and train each model on an Nvidia GeForce GTX 1080 GPU while the self-supervised examples are created in parallel using 8 processes on an Intel Core i7-7700K CPU. The code is available at \url{https://github.com/hmsch/natural-synthetic-anomalies}.

\subsection{Results and evaluation}\label{sec:results}

\noindent\textbf{Defect detection.} In \cref{tab:mvtec_image_auroc} we compare the detection performance of our models trained using variations of \placeholder{}, FPI \cite{tan2020detecting}, PII \cite{PII_miccai}, and CutPaste \cite{li2021cutpaste} to CutPaste (3-way) from \cite{li2021cutpaste} which was previously the top-performing approach for defection detection in the MVTec AD dataset \cite{BergmannPaul2021TMAD} without using additional datasets. For \placeholder{}, we report the mean and standard error of the detection AUROC for each class as well as the object, texture, and overall averages across five different random seeds. Our best method, \placeholder{} (logistic), achieves an overall image-level AUROC of \textbf{97.2} outperforming CutPaste (3-way) \cite{li2021cutpaste} by 2.0 which is well outside of the standard error range. A single \placeholder{} (logistic) model is also better than an ensemble of 5 CutPaste (3-way) models \cite{li2021cutpaste} (96.1 AUROC) and comparable to EfficientNet \cite{pmlr-v97-tan19a} pre-trained with ImageNet and finetuned with CutPaste (3-way) \cite{li2021cutpaste} (97.1 AUROC).\par 
Methods using pre-trained ImageNet models, such as PaDiM \cite{defard2020padim} (97.9 AUROC), do not provide a fair comparison to our from-scratch approach.  DRAEM \cite{zavrtanik2021draem} is a more relevant competitor. It uses the describable textures dataset (DTD) \cite{6909856} not for pre-training but for creating its synthetic anomalies. DRAEM's training objective consists of localizing and correcting the synthetic anomalies, while our model performs anomaly localization. DRAEM (98.0 AUROC) outperforms our approach slightly overall, however both approaches give comparable results for many classes despite the fact that our models only have around 11 million parameters while DRAEM's two components add up to over 97 million parameters. We note that no standard errors were reported for DRAEM's results. The authors of DRAEM claim that realism of the synthetic anomalies is not important for their method. However, their own ablation study indicates that using plain solid colors instead of real textures from DTD, causes localization AUROC and AP to drop from 97.1 and 68.4 down to 92.6 and 56.5, respectively \cite{zavrtanik2021draem}. As such, external data from DTD, which is designed to span a range of textures found in the wild \cite{6909856}, may help to produce synthetic anomalies that overlap more with real anomalies. Furthermore, when using a similar training setup as ours, \emph{i.e.}, without its reconstructive sub-network, DRAEM performs worse (93.9 AUROC) than NSA which uses more realistic synthetic anomalies.

\begin{table}[htb]
\resizebox{1.0\linewidth}{!}{%
\begin{tabular}{llcccccccc}
\toprule
& & \multicolumn{2}{c}{SOTA} & \multicolumn{6}{c}{Our Experiments} \\ \cmidrule(r{2pt}){3-4}\cmidrule(l{2pt}){5-10}
       &     & \begin{tabular}{@{}c@{}} DRAEM \\ \cite{zavrtanik2021draem} \end{tabular}  &  \begin{tabular}{@{}c@{}} CutPaste \\ (3-way) \cite{li2021cutpaste} \end{tabular}  &  \begin{tabular}{@{}c@{}} CutPaste \\ (end-to-end)  \end{tabular} &  FPI  &  PII & \begin{tabular}{@{}c@{}} \placeholder{} \\ (binary)  \end{tabular} &    \begin{tabular}{@{}c@{}} \placeholder{} \\ (continuous) \end{tabular} & \begin{tabular}{@{}c@{}} \placeholder{} \\ (logistic) \end{tabular}  \\
\midrule
\multirow{11}{*}{\rotatebox[origin=c]{90}{object}}  & bottle &   \textbf{99.2} &    98.3 \scalebox{0.7}{ $\pm$ 0.5} &                  100.0 & 90.2 &         97.6 &   97.6 \scalebox{0.7}{ $\pm$ 0.2} &   97.5 \scalebox{0.7}{ $\pm$ 0.2} &     97.7 \scalebox{0.7}{ $\pm$ 0.3} \\
      & cable &   91.8 &    80.6 \scalebox{0.7}{ $\pm$ 0.5} &                   75.4 & 68.0 &         68.9 &   92.1 \scalebox{0.7}{ $\pm$ 2.4} &   90.2 \scalebox{0.7}{ $\pm$ 3.0} &      \textbf{94.5} \scalebox{0.7}{ $\pm$ 1.0} \\
      & capsule & \textbf{98.5} &      96.2 \scalebox{0.7}{ $\pm$ 0.5} &                   89.2 & 87.5 &         84.9 &   93.2 \scalebox{0.7}{ $\pm$ 0.8} &   92.8 \scalebox{0.7}{ $\pm$ 2.2} &      95.2 \scalebox{0.7}{ $\pm$ 1.7} \\
      & hazelnut &    \textbf{100.0} &    97.3 \scalebox{0.7}{ $\pm$ 0.3} &                   81.4 & 86.0 &         82.7 &   93.5 \scalebox{0.7}{ $\pm$ 1.9} &   89.3 \scalebox{0.7}{ $\pm$ 4.9} &     94.7 \scalebox{0.7}{ $\pm$ 1.1} \\
      & metal nut &   98.7 &     \textbf{99.3} \scalebox{0.7}{ $\pm$ 0.2} &                   70.6 & 88.4 &         98.9 &    \textbf{99.4} \scalebox{0.7}{ $\pm$ 0.3} &   94.6 \scalebox{0.7}{ $\pm$ 2.1} &     98.7 \scalebox{0.7}{ $\pm$ 0.7} \\
      & pill &   \textbf{98.9} &    92.4 \scalebox{0.7}{ $\pm$ 1.3} &                   90.3 & 71.8 &         86.3 &   97.0 \scalebox{0.7}{ $\pm$ 0.9} &   94.3 \scalebox{0.7}{ $\pm$ 1.1} &      \textbf{99.2} \scalebox{0.7}{ $\pm$ 0.6} \\
      & screw &   \textbf{93.9} &    86.3 \scalebox{0.7}{ $\pm$ 1.0} &                   65.5 & 61.2 &         74.7 &   90.3 \scalebox{0.7}{ $\pm$ 1.2} &    90.1 \scalebox{0.7}{ $\pm$ 0.9} &     90.2 \scalebox{0.7}{ $\pm$ 1.4} \\
      & toothbrush &  \textbf{100.0} &     98.3 \scalebox{0.7}{ $\pm$ 0.9} &                   96.7 & 85.8 &         93.1 &   \textbf{100.0} \scalebox{0.7}{ $\pm$ 0.0} &    \textbf{99.6} \scalebox{0.7}{ $\pm$ 0.5} &     \textbf{100.0} \scalebox{0.7}{ $\pm$ 0.0} \\
      & transistor &  93.1 &      \textbf{95.5} \scalebox{0.7}{ $\pm$ 0.5} &                   88.2 & 79.6 &         90.1 &   93.5 \scalebox{0.7}{ $\pm$ 0.9} &   92.8 \scalebox{0.7}{ $\pm$ 2.2} &      \textbf{95.1} \scalebox{0.7}{ $\pm$ 0.2} \\
      & zipper &  \textbf{100.0} &     99.4 \scalebox{0.7}{ $\pm$ 0.2} &                   98.7 & 97.7 &         99.8 &   99.8 \scalebox{0.7}{ $\pm$ 0.1} &    \textbf{99.5} \scalebox{0.7}{ $\pm$ 0.7} &      99.8 \scalebox{0.7}{ $\pm$ 0.1} \\ \cmidrule(l){2-10}
      & average &   \textbf{97.4} &    94.3 \scalebox{0.7}{ $\pm$ 0.6} &                   85.6 & 81.6 &         87.7 &   95.6 \scalebox{0.7}{ $\pm$ 0.5} &   94.1 \scalebox{0.7}{ $\pm$ 1.0} &      96.5 \scalebox{0.7}{ $\pm$ 0.3} \\ \midrule
\multirow{6}{*}{\rotatebox[origin=c]{90}{texture}}  & carpet &   \textbf{97.0} &    93.1 \scalebox{0.7}{ $\pm$ 1.1} &                   53.1 & 56.0 &         65.6 &   85.6 \scalebox{0.7}{ $\pm$ 7.6} &   90.9 \scalebox{0.7}{ $\pm$ 2.2} &      95.6 \scalebox{0.7}{ $\pm$ 0.6} \\
      & grid &   \textbf{99.9} &     \textbf{99.9} \scalebox{0.7}{$\pm$ 0.1} &                   99.7 & 99.5 &        100.0 &   \textbf{99.9} \scalebox{0.7}{ $\pm$ 0.1} &   98.5 \scalebox{0.7}{ $\pm$ 3.3} &      \textbf{99.9} \scalebox{0.7}{ $\pm$ 0.1} \\
      & leather &    \textbf{100.0} &   \textbf{100.0} \scalebox{0.7}{ $\pm$ 0.0} &                   86.6 & 91.7 &        100.0 &    \textbf{99.9} \scalebox{0.7}{ $\pm$ 0.1} &   \textbf{100.0} \scalebox{0.7}{ $\pm$ 0.0} &      \textbf{99.9} \scalebox{0.7}{ $\pm$ 0.1} \\
      & tile &   99.6 &   93.4 \scalebox{0.7}{ $\pm$ 1.0} &                   87.8 & 90.2 &         98.4 &   99.7 \scalebox{0.7}{ $\pm$ 0.2} &   \textbf{100.0} \scalebox{0.7}{ $\pm$ 0.0} &     \textbf{100.0} \scalebox{0.7}{ $\pm$ 0.0} \\
      & wood &   \textbf{99.1} &     \textbf{98.6} \scalebox{0.7}{ $\pm$ 0.5} &                   84.6 & 74.4 &         91.9 &   96.7 \scalebox{0.7}{ $\pm$ 1.2} &    97.8 \scalebox{0.7}{ $\pm$ 0.8} &      97.5 \scalebox{0.7}{ $\pm$ 1.5} \\ \cmidrule(l){2-10}
      & average &  \textbf{99.1} &     97.0 \scalebox{0.7}{ $\pm$ 0.5} &                   82.4 & 82.4 &         91.2 &   96.4 \scalebox{0.7}{ $\pm$ 1.4} &   97.5 \scalebox{0.7}{ $\pm$ 0.9} &      98.6 \scalebox{0.7}{ $\pm$ 0.3} \\ \midrule
\multicolumn{2}{c}{overall average}  &   \textbf{98.0} &    95.2 \scalebox{0.7}{ $\pm$ 0.6} &                   84.5 & 81.9 &         88.9 &   95.9 \scalebox{0.7}{ $\pm$ 0.7} &   95.2 \scalebox{0.7}{ $\pm$ 0.5} &      97.2 \scalebox{0.7}{ $\pm$ 0.3} \\
\bottomrule

\end{tabular}}
\caption{Image-level AUROC \% for MVTec AD and standard error across five different random seeds. For our models, the image-level score is the average pixel score across the image. Best scores between DRAEM \cite{zavrtanik2021draem}, CutPaste (3-way) \cite{li2021cutpaste}, and \placeholder{} within standard error are bold-faced. Note that DRAEM uses additional data.}
    \label{tab:mvtec_image_auroc}
\end{table}

\noindent\textit{Synthetic anomalies should be as diverse and realistic as possible.} In our experiments, models trained with self-supervised examples created using Poisson blending clearly outperform models trained with simpler data-augmentation strategies like CutPaste \cite{li2021cutpaste} and FPI \cite{tan2020detecting}. Examples created with Poisson blending are more visually similar to real-world defects as they do not have artificial discontinuities (\cref{fig:self_sup_examples}) and according to \cref{tab:mvtec_image_auroc}, the corresponding models generalize better to real defects. 
Although PII \cite{PII_miccai} performs well for most texture classes, \placeholder{}, which also shifts and resizes the patches, performs much better for objects. Since PII and FPI use the same source and destination location for the patches, their synthetic anomalies are very subtle for aligned object classes and much subtler than the real defects in the MVTec AD dataset. 
In classes with lower AUROC, the synthetic training anomalies may have less similarity to the real test anomalies. These classes also tend to have higher standard error. In contrast, classes with high AUROC have low standard error. This could indicate that for these classes the self-supervised task can sensitize the network to the distribution of real anomalies reliably. It may also be possible to use the variance between random seeds to gauge the reliability of predictions.

\noindent\textit{Labels should approximate the degree of abnormality.} Aside from abnormal irregularities there can also be natural variation between and within the images of each class. When training a model with binary labels, the final activations tend to saturate and the predictions do not give any measure of how anomalous the regions with high scores are. Training a model with continuous labels teaches the model to differentiate between different degrees of anomalies. When using unbounded continuous labels, training is less stable and the AUROC scores have a high standard error. Models trained with bounded continuous labels outperform the binary ones most in classes with high inherent variation such as cable, hazelnut, transistor, carpet, and wood (\cref{tab:mvtec_image_auroc}) when using a threshold independent metric such as AUROC. 
PII \cite{PII_miccai} also uses continuous labels; pixels corresponding to the foreign patch are assigned to a uniform value equivalent to the interpolation factor. However, \placeholder{} (logistic) outperforms PII for some textures and most objects, including unaligned objects. For aligned objects, \placeholder{} creates more diverse anomalies than PII because the location of the source and destination patches can be different. But for unaligned objects, this advantage is negligible. Despite the similarity in generated anomalies, \placeholder{} still yields higher performance in these classes. 
A possible explanation is that the labels for \placeholder{} (logistic) are inhomogeneous and based on the outcome of the blending rather than its setup and hence more accurately represent the degree of abnormality. \par 

\begin{figure}[htb]
\includegraphics[width=\columnwidth]{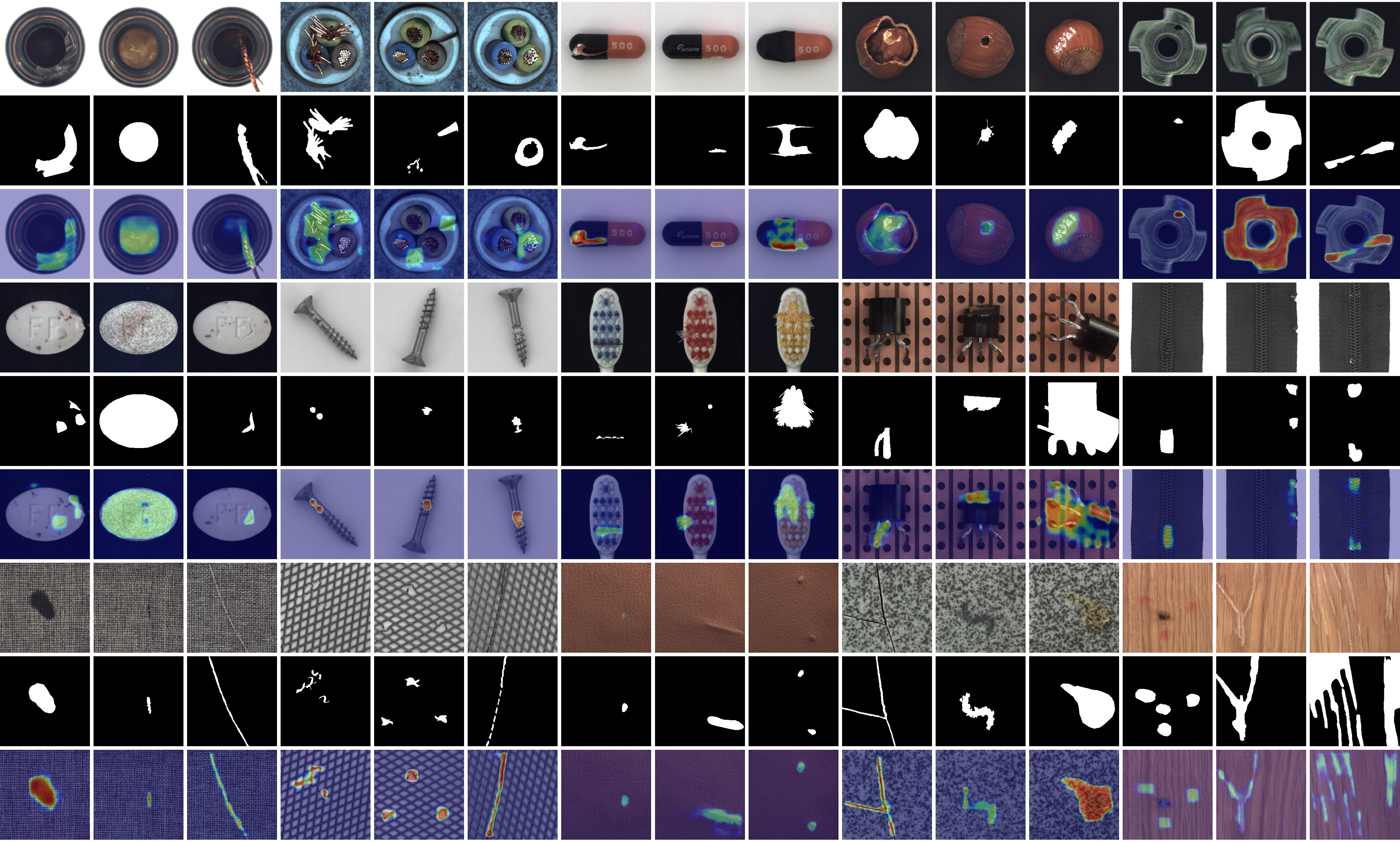}
\caption{Examples of defect localization in the MVTec AD dataset using models trained with \placeholder{} (logistic). From top to bottom: input images, human annotation, heatmap of pixel-level predictions. Best viewed in a digital version.}
\label{fig:mvtec_preds}
\end{figure}

\begin{table}[htb]
\resizebox{1.0\linewidth}{!}{%
\begin{tabular}{llcccccccc}
\toprule
& & \multicolumn{2}{c}{SOTA} & \multicolumn{6}{c}{Our Experiments} \\ \cmidrule(r{2pt}){3-4}\cmidrule(l{2pt}){5-10}
       &      & \begin{tabular}{@{}c@{}} DRAEM \\ \cite{zavrtanik2021draem} \end{tabular} &  \begin{tabular}{@{}c@{}} CutPaste \\ (3-way) \cite{li2021cutpaste} \end{tabular}  &  \begin{tabular}{@{}c@{}} CutPaste \\ (end-to-end)  \end{tabular} &  FPI  &  PII & \begin{tabular}{@{}c@{}} \placeholder{} \\ (binary)  \end{tabular} &    \begin{tabular}{@{}c@{}} \placeholder{} \\ (continuous) \end{tabular} & \begin{tabular}{@{}c@{}} \placeholder{} \\ (logistic) \end{tabular}  \\
\midrule
\multirow{11}{*}{\rotatebox[origin=c]{90}{object}} & bottle &                          \textbf{99.1} &                   97.6 \scalebox{0.7}{ $\pm$ 0.1} &                   97.7 & 91.8 &         93.1 &  98.4 \scalebox{0.7}{ $\pm$ 0.2} &  97.3 \scalebox{0.7}{ $\pm$ 0.5} &     98.3 \scalebox{0.7}{ $\pm$ 0.1} \\
      & cable &                          \textbf{94.7} &                   90.0 \scalebox{0.7}{ $\pm$ 0.2} &                   81.0 & 66.5 &         70.2 &  \textbf{93.3} \scalebox{0.7}{ $\pm$ 3.4} &  91.0 \scalebox{0.7}{ $\pm$ 2.8} &     \textbf{96.0} \scalebox{0.7}{ $\pm$ 1.4} \\
      & capsule &                          94.3 &                   97.4 \scalebox{0.7}{ $\pm$ 0.1} &                   97.5 & 95.9 &         90.2 &  \textbf{98.1} \scalebox{0.7}{ $\pm$ 0.2} &  91.6 \scalebox{0.7}{ $\pm$ 5.6} &    \textbf{97.6} \scalebox{0.7}{ $\pm$ 0.9} \\
      & hazelnut &                          \textbf{99.7} &                   97.3 \scalebox{0.7}{ $\pm$ 0.1} &                   94.8 & 89.8 &         97.0 &  97.2 \scalebox{0.7}{ $\pm$ 0.6} & 97.7 \scalebox{0.7}{ $\pm$ 0.6} &     97.6 \scalebox{0.7}{ $\pm$ 0.6} \\
      & metal nut &                          \textbf{99.5} &                   93.1 \scalebox{0.7}{ $\pm$ 0.4} &                   68.1 & 96.2 &         95.4 &  98.2 \scalebox{0.7}{ $\pm$ 0.2} &  97.3 \scalebox{0.7}{ $\pm$ 0.3} &     \textbf{98.4} \scalebox{0.7}{ $\pm$ 0.2} \\
      & pill &                          97.6 &                   95.7 \scalebox{0.7}{ $\pm$ 0.1} &                   98.1 & 62.3 &         95.3 &  \textbf{98.5} \scalebox{0.7}{ $\pm$ 0.2} &  \textbf{97.1} \scalebox{0.7}{ $\pm$ 2.7} &     \textbf{98.5} \scalebox{0.7}{ $\pm$ 0.3} \\
      & screw &                          \textbf{97.6} &                   96.7 \scalebox{0.7}{ $\pm$ 0.1} &                   90.7 & 90.4 &         92.8 &  96.7 \scalebox{0.7}{ $\pm$ 0.4} &  \textbf{92.3} \scalebox{0.7}{ $\pm$ 5.3} &     96.5 \scalebox{0.7}{ $\pm$ 0.1} \\
      & toothbrush &                          \textbf{98.1} &                   \textbf{98.1} \scalebox{0.7}{ $\pm$ 0.0} &                   95.7 & 81.8 &         81.3 &  95.6 \scalebox{0.7}{ $\pm$ 0.6} &  94.5 \scalebox{0.7}{ $\pm$ 0.7} &     94.9 \scalebox{0.7}{ $\pm$ 0.7} \\
      & transistor &                          90.9 &                   \textbf{93.0} \scalebox{0.7}{ $\pm$ 0.2} &                   85.9 & 78.5 &         86.9 &  87.8 \scalebox{0.7}{ $\pm$ 1.9} &  80.2 \scalebox{0.7}{ $\pm$ 3.3} &     88.0 \scalebox{0.7}{ $\pm$ 1.8} \\
      & zipper &                          98.8 &                   \textbf{99.3} \scalebox{0.7}{ $\pm$ 0.0} &                   92.9 & 91.8 &         93.8 &  94.2 \scalebox{0.7}{ $\pm$ 0.2} &  90.7 \scalebox{0.7}{ $\pm$ 1.5} &     94.2 \scalebox{0.7}{ $\pm$ 0.3} \\ \cmidrule(l){2-10}
      & average &                          \textbf{97.0} &                   95.8 \scalebox{0.7}{ $\pm$ 0.1} &                   90.2 & 84.5 &         89.6 &  95.8 \scalebox{0.7}{ $\pm$ 0.4} &  93.0 \scalebox{0.7}{ $\pm$ 1.7} &     96.0 \scalebox{0.7}{ $\pm$ 0.4} \\ \midrule
\multirow{6}{*}{\rotatebox[origin=c]{90}{texture}} & carpet &                          95.5 &                   \textbf{98.3} \scalebox{0.7}{ $\pm$ 0.0} &                   83.3 & 70.8 &         97.2 &  \textbf{94.5} \scalebox{0.7}{ $\pm$ 4.1} &  81.8 \scalebox{0.7}{ $\pm$ 6.8} &     95.5 \scalebox{0.7}{ $\pm$ 2.3} \\
      & grid &                          \textbf{99.7} &                   97.5 \scalebox{0.7}{ $\pm$ 0.1} &                   97.6 & 94.2 &         98.9 &  99.1 \scalebox{0.7}{ $\pm$ 0.0} &  98.0 \scalebox{0.7}{ $\pm$ 0.3} &     99.2 \scalebox{0.7}{ $\pm$ 0.1} \\
      & leather &                          98.6 &                   \textbf{99.5} \scalebox{0.7}{ $\pm$ 0.0} &                   96.4 & 88.3 &         99.2 &  \textbf{99.6} \scalebox{0.7}{ $\pm$ 0.0} &  \textbf{99.5} \scalebox{0.7}{ $\pm$ 0.2} &     \textbf{99.5} \scalebox{0.7}{ $\pm$ 0.1} \\
      & tile &                          99.2 &                   90.5 \scalebox{0.7}{ $\pm$ 0.2} &                   72.7 & 65.0 &         98.0 &  99.0 \scalebox{0.7}{ $\pm$ 0.2} &  97.4 \scalebox{0.7}{ $\pm$ 0.7} &     \textbf{99.3} \scalebox{0.7}{ $\pm$ 0.0} \\
      & wood &                          \textbf{96.4} &                   95.5 \scalebox{0.7}{ $\pm$ 0.1} &                   84.0 & 71.1 &         91.1 &  94.0 \scalebox{0.7}{ $\pm$ 0.8} &  90.6 \scalebox{0.7}{ $\pm$ 3.7} &     90.7 \scalebox{0.7}{ $\pm$ 1.9} \\ \cmidrule(l){2-10}
      & average &                          \textbf{97.9} &                   96.3 \scalebox{0.7}{ $\pm$ 0.1} &                   86.8 & 77.9 &         96.9 &  97.3 \scalebox{0.7}{ $\pm$ 0.7} &  93.5 \scalebox{0.7}{ $\pm$ 0.9} &     96.8 \scalebox{0.7}{ $\pm$ 0.7} \\ \midrule
\multicolumn{2}{c}{overall average} &                          \textbf{97.3} &                   96.0 \scalebox{0.7}{ $\pm$ 0.1} &                   89.1 & 82.3 &         92.0 &  96.3 \scalebox{0.7}{ $\pm$ 0.2} &  93.1 \scalebox{0.7}{ $\pm$ 1.1} &     96.3 \scalebox{0.7}{ $\pm$ 0.4} \\
\bottomrule
\end{tabular}}
  \caption{Pixel-level AUROC \% with standard error for the MVTec AD models from \cref{tab:mvtec_image_auroc}. Note that DRAEM
uses additional data.}
    \label{tab:mvtec_pixel_auroc}
\end{table}
\noindent\textbf{Defect localization.} In \cref{tab:mvtec_pixel_auroc} we report the pixel-level performance of the models from \cref{tab:mvtec_image_auroc}. Although \placeholder{} performs well for objects on the image-level when trained with unbounded continuous labels, pixel-wise performance is much better for \placeholder{} with bounded binary or continuous labels. \placeholder{} (logistic) achieves a \textbf{96.3} average pixel-level AUROC performing similarly to CutPaste (3-way) \cite{li2021cutpaste} (96.0 AUROC). It also reaches similar performance to DRAEM \cite{zavrtanik2021draem} for many classes although DRAEM achieves a higher overall score (97.3 AUROC). \cref{fig:mvtec_preds} shows that \placeholder{} (logistic) can localize a wide range of real-world defects accurately including anomalies that are very different from the synthetic anomalies seen during training (\emph{e.g.}, white writing on hazelnuts, misplaced transistors, stained tiles and carpet). 

\noindent\textbf{Medical imaging.} 
\begin{table}[htb]
\begin{tabular}{lcccccc}
\toprule
& \begin{tabular}{@{}c@{}} CutPaste \\ (end-to-end)  \end{tabular} &  FPI  &  PII & \begin{tabular}{@{}c@{}} \placeholder \\ (binary)  \end{tabular} &    \begin{tabular}{@{}c@{}} \placeholder \\ (continuous) \end{tabular} & \begin{tabular}{@{}c@{}} \placeholder \\ (logistic) \end{tabular}  \\
\midrule
male &      59.8 & 73.7 &  91.7 \scalebox{0.7}{ $\pm$ 0.6} &  \textbf{94.0} \scalebox{0.7}{ $\pm$ 0.5} &  \textbf{93.4} \scalebox{0.7}{ $\pm$ 0.3} &  \textbf{94.0} \scalebox{0.7}{ $\pm$ 0.6} \\
female &      56.2 & 67.4 &  92.8 \scalebox{0.7}{ $\pm$ 0.4} &  \textbf{94.3} \scalebox{0.7}{ $\pm$ 0.6} &  93.0 \scalebox{0.7}{ $\pm$ 0.4} &  \textbf{94.0} \scalebox{0.7}{ $\pm$ 0.5} \\
\bottomrule
\end{tabular}
\caption{Image-level AUROC \% for rCXR and standard error across five different random seeds. Best scores per row within standard error are bold-faced.}
    \label{tab:rCXR}
\end{table}
\begin{figure}[htb]
\includegraphics[width=1.0\columnwidth]{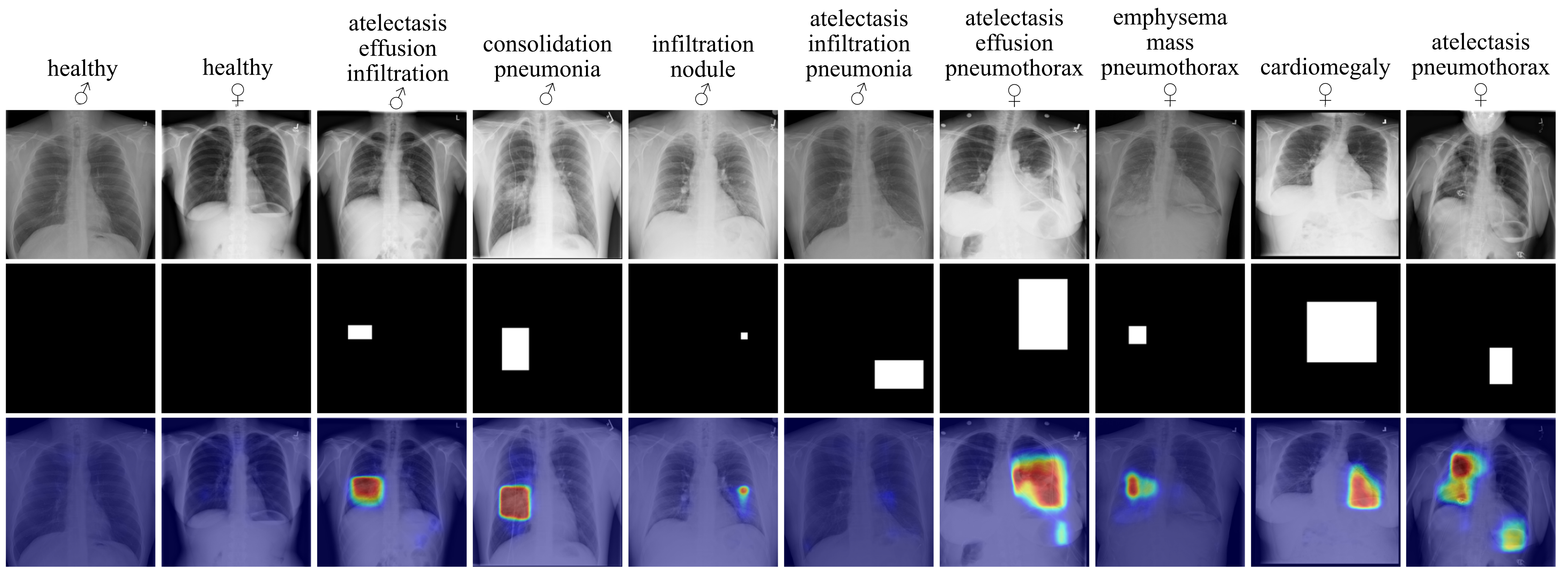}
\caption{Example localization predictions for chest X-ray disease detection using models trained with \placeholder{} (binary). From top to bottom: input images, rough bounding box from a radiologist, heatmap of pixel-level predictions. Each case is labeled with pathology keywords that \cite{wang2017chestx} mined from radiologist reports. }
\label{fig:rCXR_preds}
\end{figure}
In \cref{tab:rCXR}, we compare the performance of \placeholder{} and end-to-end models trained using other self-supervised tasks for the task of binary classification of rCXR images into healthy (normal) and pathological (abnormal) categories. Models trained using \placeholder{} clearly outperform end-to-end models trained with FPI \cite{tan2020detecting} or CutPaste \cite{li2021cutpaste}. \placeholder{} also outperforms PII \cite{PII_miccai}. However, the type of label used for \placeholder{} is less important for this dataset. Since there is high inter-sample variability in the normal data, it is possible that all synthetic anomalies created by \placeholder{} would be considered abnormal. So, approximating the degree of abnormality with a continuous label does not improve over the binary labels when evaluating the model with real anomalies. \par
We do not report pixel-level metrics, as we only have very rough bounding boxes for less than 10\% of the test set. \cref{fig:rCXR_preds} shows example predictions for several healthy and pathological cases. The localization predictions are good for examples 1--5, 7, and 8, but the model fails to detect any abnormal findings in the 6th example and disagrees with the bounding box annotation for bottom examples 9 and 10. 
We did not compare to CutPaste (3-way)~\cite{li2021cutpaste}  here as there is no reference implementation available.

\noindent\textbf{Limitations.}  The 9th example in \cref{fig:rCXR_preds} shows cardiomegaly, \emph{i.e.}, an enlarged heart, for which the bounding box contains the entire heart. But the model only activates for portions of the heart that exceed the normal size found in healthy patients. In these cases, the model lacks the semantic understanding that radiologists use to categorize diseases.
In the 10th example, the model activates more for the tubes on the patient's right side than for the finding inside of the bounding box. Unlike a human radiologist, an anomaly detection model cannot be expected to automatically classify clinically correctly placed lines, tubes, or cardiac devices as normal if they are not expected in the healthy training distribution.\par
\begin{figure}[htb]
  \includegraphics[width=1.0\linewidth]{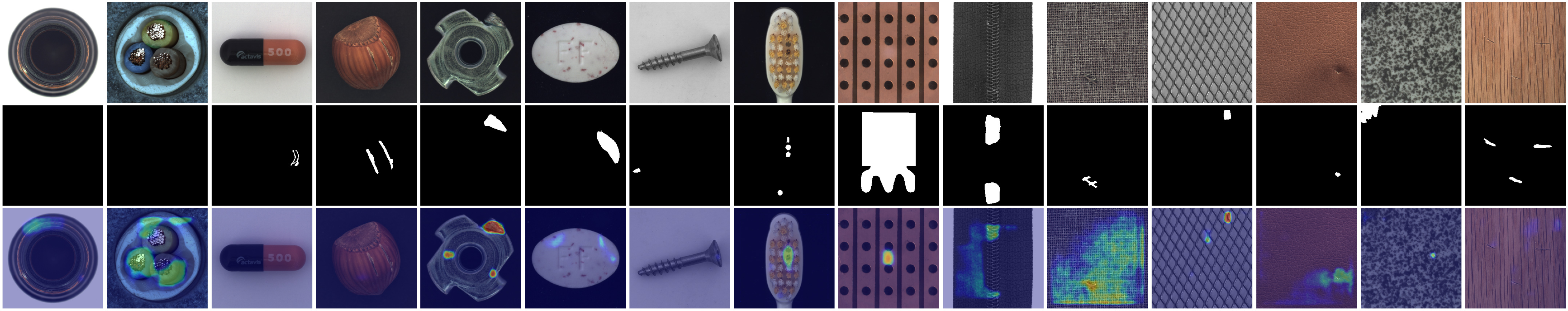}
  \caption{Failure cases for MVTec AD defect localization using models trained with \placeholder{} (logistic). Examples include false pos./negatives and incorrect localization. From left to right: input, human annotation, heatmap of pixel-level predictions.}
  \label{fig:mvtec_failures}
\end{figure}

Our method sometimes fails to detect very small defects (see capsule, hazelnut, screw, and wood examples in \cref{fig:mvtec_failures}), predicts too large regions or has false positives (see bottle, cable, zipper, carpet, leather in \cref{fig:mvtec_failures}). The transistor class has several examples of misplaced or missing transistors. These are detected but the predicted localization does not match the large human annotation (\cref{fig:mvtec_failures}) resulting in a low localization AUROC (\cref{tab:mvtec_pixel_auroc}). Patch-wise localization, as used for CutPaste \cite{li2021cutpaste} and PaDiM \cite{defard2020padim} among others, can detect the missing transistor for each patch and hence produce a segmentation map closer to the human annotation. However, for these large anomalies the type of defect is immediately obvious to a human inspector once detected so we argue that precise localization would not be necessary for most applications. \par 
Since the model does not see any real anomalies during training, it may predict statistically unlikely normal variations as abnormal and fails to recognize subtle anomalies that are very different from the synthetic anomalies seen during training. Hence, predictions from a self-supervised anomaly detection model should not be used on their own for  decision making. Such models can however be useful as an instant second observer and for quality control. As far as we are aware, there are no further potential negative societal impacts of this work. 

\section{Conclusion}
We propose a self-supervised task that creates diverse and realistic synthetic anomalies. These training examples are generated under controlled conditions that help to produce relevant and subtle anomalies. This provides a more consistent training signal and results in better detection of real anomalies. The formulation of the loss and synthetic labels yields an effective and computationally efficient training task. This helps \placeholder{} outperform state-of-the-art methods on both natural and medical imaging datasets, demonstrating its generalizability. In the future, additions such as quantifying uncertainty or exploiting classes of known anomalies could help facilitate the use of \placeholder{} in  critical applications.

\noindent\textbf{Acknowledgments.} This work was supported by the UK Research and Innovation London Medical Imaging and Artificial Intelligence Centre for Value Based Healthcare and the iFind project, Wellcome Trust IEH Award [102431].

\clearpage
%
%
\bibliographystyle{splncs04}
\bibliography{references}

\clearpage
\appendix
\section{Additional experiment details}
\subsection{Hyperparameters}
The hyperparameters for the self-supervised tasks used in our experiments are given in \cref{tab:hyperparams}. $n_\text{max}$ is the maximum number of patches added to each image. $h_{\text{min}}, h_{\text{max}}, w_{\text{min}}, w_{\text{max}} \in [0, 1]$ are the bounds on the patch dimensions relative to the image size. $b \in [0, 225]$ is the background brightness. All pixels with absolute brightness distance less than $t_\text{brightness}$ to the background brightness are assigned to the background. $t_\text{object}, t_\text{overlap} \in [0,1]$ are used for the object and overlap conditions for the patches (see Section \ref{sec:nsa}). The patch resize-scale is clipped to the range $[s_\text{min}, s_\text{max}]$ in addition to the conditions given in Section \ref{sec:nsa}. $y_0$ and $k$ give the midpoint and steepness of the logistic function used for creating the labels for \placeholder{} (logistic). \par 
These parameters encode our assumptions about the unknown real out-distribution (see \cref{sec:nsa}). Thus they were not tuned in a data-driven way as no validation set containing all possible types of real anomalies was available. These assumptions were chosen based on visual inspection of the input images and self-supervised examples. \emph{E.g.}, for objects that have larger width than height, $w_{\text{max}}$ is higher than $h_{\text{max}}$ and vice versa; for classes with high perceived natural variation $y_0$ should be larger and $k$ smaller. \par 
 
\begin{table}[H]
\caption{Hyperparameters for the self-supervised tasks.}
    \label{tab:hyperparams}
    \centering
\resizebox{1.0\linewidth}{!}{%
\begin{tabular}{lllcccccccccc}
\toprule
&&&&\multicolumn{2}{c}{patch size}&\multicolumn{4}{c}{background constraints}&\multicolumn{1}{c}{scale}&\multicolumn{2}{c}{logistic}\\\cmidrule(l{2pt}r{2pt}){5-6}\cmidrule(l{2pt}r{2pt}){7-10}\cmidrule(l{2pt}r{2pt}){11-11}\cmidrule(l{2pt}){12-13}
                           &                          &            & $n_\text{max}$ & $h_{\text{min}}, h_{\text{max}}$ & $w_{\text{min}}, w_{\text{max}}$ & $b$ & $t_\text{brightness}$ &
                           $t_\text{object}$ &
                           $t_\text{overlap}$ &
                           $s_\text{min}, s_\text{max}$ & $y_0$ & $k$    \\ \midrule
\multirow{15}{*}{\rotatebox[origin=c]{0}{MVTec AD}} & \multirow{10}{*}{\rotatebox[origin=c]{90}{object}} & bottle     & 3              & 0.06,  0.80                      & 0.06,  0.80                      & 200 & 60    & 0.70 & 0.25 & 0.7, 1.3                 & 24    & $1/12$ \\
                           &                          & cable      & 3              & 0.10,  0.80                      & 0.10,  0.80                      & N/A & N/A  &  N/A & N/A & 0.7, 1.3                 & 24    & $1/12$ \\
                           &                          & capsule    & 3              & 0.06,  0.30                      & 0.06,  0.80                      & 200 & 60 &  0.70 & 0.25 & 0.7, 1.3                   & 4     & $1/2$  \\
                           &                          & hazelnut   & 3              & 0.06,  0.70                      & 0.06,  0.70                      & 20 & 20 &  0.70 & 0.25  & 0.7, 1.3                  & 24    & $1/12$ \\
                           &                          & metal nut  & 3              & 0.06,  0.80                      & 0.06,  0.80                      & 20  & 20 &  0.50 & 0.25  & 0.7, 1.3                   & 7     & $1/3$  \\
                           &                          & pill       & 3              & 0.06,  0.40                      & 0.06,  0.80                      & 20  & 20 &  0.70 & 0.25  & 0.7, 1.3                  & 7     & $1/3$  \\
                           &                          & screw      & 4              & 0.06,  0.24                      & 0.06,  0.24                      & 200 & 60 &  0.50 & 0.25 & 0.7, 1.3                   & 3     & $1$    \\
                           &                          & toothbrush & 3              & 0.06,  0.80                      & 0.06,  0.40                      & 20  & 20 &  0.25 & 0.25 & 0.7, 1.3                   & 15    & $1/6$  \\
                           &                          & transistor & 3              & 0.06,  0.80                      & 0.06,  0.80                      & N/A & N/A &  N/A & N/A & 0.7, 1.3                  & 15    & $1/6$  \\
                           &                          & zipper     & 4              & 0.06,  0.80                      & 0.06,  0.40                      & 200 & 60 &  0.70 & 0.25  & 0.7, 1.3                  & 15    & $1/6$  \\ \cmidrule(l){2-13}
                           & \multirow{5}{*}{\rotatebox[origin=c]{90}{texture}} & carpet     & 4              & 0.06,  0.80                      & 0.06,  0.80        & N/A & N/A   &  N/A & N/A & 0.5, 2.0               & 7     & $1/3$  \\
                           &                          & grid       & 4              & 0.06,  0.80                      & 0.06,  0.80                      & N/A & N/A & N/A & N/A & 0.5, 2.0                      & 7     & $1/3$  \\
                           &                          & leather    & 4              & 0.06,  0.80                      & 0.06,  0.80                      & N/A & N/A & N/A & N/A   & 0.5, 2.0                   & 7     & $1/3$  \\
                           &                          & tile       & 4              & 0.06,  0.80                      & 0.06,  0.80                      & N/A & N/A & N/A & N/A & 0.5, 2.0                     & 7     & $1/3$  \\
                           &                          & wood       & 4              & 0.06,  0.80                      & 0.06,  0.80                      & N/A & N/A & N/A & N/A  & 0.5, 2.0                   & 15    & $1/6$  \\ \midrule
\multirow{2}{*}{\rotatebox[origin=c]{0}{rCXR}}      & \multicolumn{2}{l}{male}              & 3              & 0.06,  0.80                      & 0.06,  0.80                      & 0   & 20 &  0.70 & 0.70   & 0.7, 1.3                 & 4     & $1/2$  \\
                           & \multicolumn{2}{l}{female}            & 3              & 0.06,  0.80                      & 0.06,  0.80                      & 0   & 20  &  0.70 & 0.70   & 0.7, 1.3              & 4     & $1/2$ \\ \bottomrule
\end{tabular}}
\end{table}

For FPI (Poisson), we used mixed gradients for seamless cloning. For \placeholder{}, we use mixed gradients for all rCXR data and for MVTec AD texture classes. For MVTec AD object classes, we find that OpenCV's \cite{opencv_library} seamless cloning method causes artifacts when there are sharp contrast changes (\emph{e.g.}, at the boundary from the object to the background) near the edges of the patch boundary more frequently when using mixed gradients than source gradients. Thus, we only use source gradients for these classes for \placeholder{}. \par

\subsection{Comparison of self-supervised tasks}
\begin{table}[H]
\caption{Comparison of CutPaste \cite{li2021cutpaste}, FPI \cite{tan2020detecting}, PII \cite{PII_miccai}, and NSA self-supervised tasks.}
    \label{tab:selfsup}
    \centering

\resizebox{0.98\columnwidth}{!}{%
\begin{tabular}{@{}p{3.5cm}p{2.3cm}p{2.3cm}p{2.3cm}p{2.3cm}p{2.3cm}p{2.3cm}@{}}
\toprule
                                                      & \hfil\begin{tabular}[c]{@{}c@{}}CutPaste  \cite{li2021cutpaste}\end{tabular}    & \hfil\begin{tabular}[c]{@{}c@{}}FPI \cite{tan2020detecting}\end{tabular}                                        & \hfil\begin{tabular}[c]{@{}c@{}}PII\cite{PII_miccai}\end{tabular} & \hfil\begin{tabular}[c]{@{}c@{}}NSA\\ (binary)\end{tabular} & \hfil\begin{tabular}[c]{@{}c@{}}NSA\\ (continuous)\end{tabular} & \hfil\begin{tabular}[c]{@{}c@{}}NSA\\ (logistic)    \end{tabular}                                         \\ \midrule
Different source and destination images?          & \hfil{\large\xmark}     & \hfil{\large\cmark}                                & \hfil{\large\cmark}                                              & \hfil{\large\cmark}                                             & \hfil{\large\cmark}                                                 & \hfil{\large\cmark}                                                                                    \\\midrule[0.25pt]
Resize patch before blending?                        & \hfil{\large\xmark}     &\hfil {\large\xmark}                                    & \hfil{\large\xmark}                                                  & {\hfil\large\cmark}                                             & \hfil{\large\cmark}                                                 & \hfil{\large\cmark}                                                                                \\\midrule[0.25pt]
Different source and destination patch locations? & \hfil{\large\cmark} & \hfil{\large\xmark}                                    & \hfil{\large\xmark}                                                  & \hfil{\large\cmark}                                             & \hfil{\large\cmark}                                                 & \hfil{\large\cmark}                                                                                    \\\midrule[0.25pt]
Blending mode                                        & \hfil copy-paste & \hfil\begin{tabular}[c]{@{}c@{}} linear \\ interpolation \end{tabular}                      & \hfil\begin{tabular}[c]{@{}c@{}}  seamless \\ cloning  \end{tabular}                  & \hfil\begin{tabular}[c]{@{}c@{}}  seamless \\ cloning  \end{tabular}       & \hfil\begin{tabular}[c]{@{}c@{}}  seamless \\ cloning  \end{tabular}          & \hfil\begin{tabular}[c]{@{}c@{}}  seamless \\ cloning  \end{tabular}                                     \\\midrule[0.25pt]
Label type                                           & \hfil binary     & \hfil\begin{tabular}[c]{@{}c@{}}  bounded \\ continuous \\  (interpolation \\ factor) \end{tabular}  & \hfil\begin{tabular}[c]{@{}c@{}}  bounded \\ continuous \\  (interpolation \\ factor) \end{tabular}   &  \hfil binary                                                 &  \hfil\begin{tabular}[c]{@{}c@{}}  continuous \\(difference-\\based)   \end{tabular}                &\hfil\begin{tabular}[c]{@{}c@{}}  bounded \\ continuous \\(difference-\\based)   \end{tabular} \\\midrule[0.5pt]
Loss                                                 & \hfil BCE-loss   & \hfil BCE-loss                                  & \hfil BCE-loss                                                & \hfil BCE-loss                                               & \hfil MSE-loss                                                        & \hfil BCE-loss    \\ \bottomrule                             
\end{tabular}}
\end{table}

\begin{table}[H]
\caption{Comparison of the original patch-selection procedures for CutPaste \cite{li2021cutpaste}, FPI \cite{tan2020detecting}, PII \cite{PII_miccai}, and our method. We use our patch-selection procedure for our re-implementations of CutPaste, FPI, and PII. See \cref{sec:nsa} for more details on our method.}
    \label{tab:patch_sampling}
\resizebox{0.98\columnwidth}{!}{%
\begin{tabular}{@{}p{3.5cm}p{5cm}p{5cm}p{5cm}@{}}
\toprule
                      & \hfil CutPaste \cite{li2021cutpaste}                                 &  \hfil FPI \cite{tan2020detecting} and PII \cite{PII_miccai}                                              &  \hfil Ours                                                                                              \\\midrule
Patch size            & area ratio between patch and image sampled from $(0.02, 0.15)$ & width and height relative to image dimensions sampled from $U(0.1, 0.4)$ & width and height relative to image dimensions sampled from truncated $\text{Gamma}(2, 0.1)$              \\\midrule[0.25pt]
Patch aspect ratio     & sampled from $(0.3,1) \cup (1, 3.3)$                           & square, except when truncated by the image boundary                      & any ratio resulting from the above                                                                \\\midrule[0.25pt]
Location restrictions & entire patch must appear in the full image                     & patch center must lie within the core 80\% of the image dimensions     & patch must contain part of the object and object portions at source and destination must overlap \\\bottomrule
\end{tabular}}
\end{table}

\clearpage
\section{Additional results}
\subsection{Additional ablation studies}
To validate our design choices we conducted several ablation studies beyond comparing different label definitions and comparing NSA to simpler baseline synthetic anomalies (see \cref{sec:results}). Additional variants of NSA (logistic) considered were:
\begin{itemize}
    \item[\textbf{A}] Do not use foreground constraints for any classes. Note that in the original, foreground constraints were not applied to cable, transistor, and textures so these experiments do not need to be duplicated.
    \item[\textbf{B}] Only use a single patch per training example instead of a random number of patches.
    \item[\textbf{C}] {Generate patch shapes as for CutPaste \cite{li2021cutpaste}: \begin{enumerate}
        \item sample the area ratio between the patch and the full image from $(0.02, 0.15)$,
        \item determine the aspect ratio by sampling from $(0.3, 1) \cup (1, 3.3)$,
        \item sample location such that patch is contained entirely within the image.
    \end{enumerate} 
(Use single patch, uniform distributions, no foreground constraints, no resizing.)}
    \item[\textbf{D}] Mask the patches with a union of 5 random ellipses to achieve non-rectangular patch shapes.
\end{itemize}

For these experiments we report image-level and pixel-level AUROC \% (\cref{tab:mvtec_ablation}). The results back-up our design choices as the final version outperforms all three variants. Specifically, the experiments show that

\begin{itemize}
    \item[\textbf{A}] using foreground constraints is most important for classes where the images contain a lot of background due to the shape of the objects (\emph{e.g.}, screw and capsule),
    \item[\textbf{B}] using a random number of patches performs slightly better than using a single patch, 
    \item[\textbf{C}] our patch-selection procedure leads to much better overall performance of NSA than the patch selection procedure described in \cite{li2021cutpaste}, and 
    \item[\textbf{D}] beyond the diverse sizes and aspect ratios the shape of the patches is not important. This could be due to the fact that because of Poisson blending rectangular patches due not necessarily create rectangular anomalies, so NSA with rectangular patches already creates various non-rectangular anomalies.
\end{itemize}

\begin{table}[H]
\caption{Image-level and pixel-level AUROC \% for MVTec AD and standard error across five different random seeds for NSA (logistic) variants.}
    \label{tab:mvtec_ablation}
    \centering
\resizebox{\linewidth}{!}{%
\begin{tabular}{llcccccccccc}
\toprule
& & \multicolumn{5}{c}{Image-level AUROC \%} & \multicolumn{5}{c}{Pixel-level AUROC \%} \\ \cmidrule(r{2pt}){3-7}\cmidrule(l{2pt}){8-12}
  \multicolumn{2}{c}{NSA (logistic) variants}  & final & A & B & C & D & final & A & B & C & D \\
\midrule
\multirow{11}{*}{\rotatebox[origin=c]{90}{object}} 
      & bottle &  97.7   \scalebox{0.7}{ $\pm$ 0.3} & 97.5  \scalebox{0.7}{ $\pm$ 0.5}  & 97.1  \scalebox{0.7}{ $\pm$ 0.4} & 97.1  \scalebox{0.7}{ $\pm$ 0.7} & 98.4 \scalebox{0.7}{ $\pm$ 0.2} & 98.3  \scalebox{0.7}{ $\pm$ 0.1} & 98.4  \scalebox{0.7}{ $\pm$ 0.1} & 97.9  \scalebox{0.7}{ $\pm$ 0.1} & 97.4 \scalebox{0.7}{ $\pm$ 0.2}&  98.9 \scalebox{0.7}{ $\pm$ 0.0} \\
      & cable &     94.5  \scalebox{0.7}{ $\pm$ 1.0} & -- & 96.1  \scalebox{0.7}{ $\pm$ 1.0} & 92.4  \scalebox{0.7}{ $\pm$ 2.0}& 91.8 \scalebox{0.7}{ $\pm$ 1.4}  & 96.0  \scalebox{0.7}{ $\pm$ 1.4} & --  & 94.7  \scalebox{0.7}{ $\pm$ 2.7} & 96.7  \scalebox{0.7}{ $\pm$ 0.5}&  86.8 \scalebox{0.7}{ $\pm$ 2.4} \\
      & capsule & 95.2  \scalebox{0.7}{ $\pm$ 1.7} & 91.7  \scalebox{0.7}{ $\pm$ 1.9}  & 89.7  \scalebox{0.7}{ $\pm$ 1.3} & 84.6  \scalebox{0.7}{ $\pm$ 0.8}& 95.6 \scalebox{0.7}{ $\pm$ 0.7} & 97.6  \scalebox{0.7}{ $\pm$ 0.9} & 96.4  \scalebox{0.7}{ $\pm$ 0.4} & 95.2  \scalebox{0.7}{ $\pm$ 1.4} & 92.7  \scalebox{0.7}{ $\pm$ 0.9} &  97.1 \scalebox{0.7}{ $\pm$ 0.2} \\
      & hazelnut     & 94.7  \scalebox{0.7}{ $\pm$ 1.1} & 93.1  \scalebox{0.7}{ $\pm$ 0.8}  & 92.2  \scalebox{0.7}{ $\pm$ 2.7} & 85.2  \scalebox{0.7}{ $\pm$ 2.5}  & 94.1 \scalebox{0.7}{ $\pm$ 2.0} & 97.6  \scalebox{0.7}{ $\pm$ 0.6} & 97.5  \scalebox{0.7}{ $\pm$ 0.7} & 93.6  \scalebox{0.7}{ $\pm$ 0.9} & 94.5  \scalebox{0.7}{ $\pm$ 1.2} &  97.5 \scalebox{0.7}{ $\pm$ 0.4} \\
      & metal nut  & 98.7  \scalebox{0.7}{ $\pm$ 0.7} & 99.0  \scalebox{0.7}{ $\pm$ 0.6}  & 97.7  \scalebox{0.7}{ $\pm$ 1.0} & 94.4  \scalebox{0.7}{ $\pm$ 1.0}  & 99.3 \scalebox{0.7}{ $\pm$ 0.3}& 98.4  \scalebox{0.7}{ $\pm$ 0.2} & 98.5  \scalebox{0.7}{ $\pm$ 0.1} & 96.5  \scalebox{0.7}{ $\pm$ 0.9} & 97.0 \scalebox{0.7}{ $\pm$ 0.3}&  98.2 \scalebox{0.7}{ $\pm$ 0.4} \\
      & pill &  99.2  \scalebox{0.7}{ $\pm$ 0.6} & 99.0  \scalebox{0.7}{ $\pm$ 0.2} & 97.8  \scalebox{0.7}{ $\pm$ 0.5} & 94.5  \scalebox{0.7}{ $\pm$ 1.6} &   96.9 \scalebox{0.7}{ $\pm$ 1.0}& 98.5  \scalebox{0.7}{ $\pm$ 0.3}& 97.5  \scalebox{0.7}{ $\pm$ 0.2} & 90.5  \scalebox{0.7}{ $\pm$ 4.5} & 92.8  \scalebox{0.7}{ $\pm$ 2.2}&  97.1 \scalebox{0.7}{ $\pm$ 1.0} \\
      & screw  & 90.2  \scalebox{0.7}{ $\pm$ 1.4} & 77.8  \scalebox{0.7}{ $\pm$ 3.3}  & 85.3  \scalebox{0.7}{ $\pm$ 3.4} & 56.3  \scalebox{0.7}{ $\pm$ 1.8}&  90.3 \scalebox{0.7}{ $\pm$ 1.0}  & 96.5  \scalebox{0.7}{ $\pm$ 0.1} & 92.9  \scalebox{0.7}{ $\pm$ 0.6} & 95.6  \scalebox{0.7}{ $\pm$ 0.8} & 82.6  \scalebox{0.7}{ $\pm$ 1.6} &  96.2 \scalebox{0.7}{ $\pm$ 0.2} \\
      & toothbrush &   100.0  \scalebox{0.7}{ $\pm$ 0.0} & 100.0  \scalebox{0.7}{ $\pm$ 0.0}  & 100.0  \scalebox{0.7}{ $\pm$ 0.0} & 99.7  \scalebox{0.7}{ $\pm$ 0.2} &  100.0 \scalebox{0.7}{ $\pm$ 0.0}& 94.9  \scalebox{0.7}{ $\pm$ 0.7} & 93.8  \scalebox{0.7}{ $\pm$ 1.1} & 91.7  \scalebox{0.7}{ $\pm$ 2.8} & 94.4 \scalebox{0.7}{ $\pm$ 0.9}&  95.3 \scalebox{0.7}{ $\pm$ 0.2} \\
      & transistor & 95.1  \scalebox{0.7}{ $\pm$ 0.2} & -- & 93.7  \scalebox{0.7}{ $\pm$ 1.5} & 91.2  \scalebox{0.7}{ $\pm$ 1.7}&   93.2 \scalebox{0.7}{ $\pm$ 0.8} & 88.0  \scalebox{0.7}{ $\pm$ 1.8} & -- & 83.8  \scalebox{0.7}{ $\pm$ 0.9} & 83.1  \scalebox{0.7}{ $\pm$ 2.2}&  86.0 \scalebox{0.7}{ $\pm$ 1.1} \\
      & zipper & 99.8  \scalebox{0.7}{ $\pm$ 0.1} & 100.0  \scalebox{0.7}{ $\pm$ 0.0}  & 99.8  \scalebox{0.7}{ $\pm$ 0.3} & 98.9  \scalebox{0.7}{ $\pm$ 1.1} &  99.9 \scalebox{0.7}{ $\pm$ 0.1} & 94.2  \scalebox{0.7}{ $\pm$ 0.3} & 94.1  \scalebox{0.7}{ $\pm$ 0.2} & 94.0  \scalebox{0.7}{ $\pm$ 0.3} & 94.0 \scalebox{0.7}{ $\pm$ 0.3} &  94.9 \scalebox{0.7}{ $\pm$ 0.1} \\ \cmidrule(l){2-12}
      & average & 96.5  \scalebox{0.7}{ $\pm$ 0.3} & -- & 94.9  \scalebox{0.7}{ $\pm$ 0.6} & 89.4  \scalebox{0.7}{ $\pm$ 0.4}  &  96.0 \scalebox{0.7}{ $\pm$ 0.2} & 96.0  \scalebox{0.7}{ $\pm$ 0.4} & -- & 93.3  \scalebox{0.7}{ $\pm$ 0.9} & 92.3  \scalebox{0.7}{ $\pm$ 0.5}&  94.8 \scalebox{0.7}{ $\pm$ 0.3} \\ \midrule
\multirow{6}{*}{\rotatebox[origin=c]{90}{texture}} 
      & carpet & 95.6  \scalebox{0.7}{ $\pm$ 0.6} & -- & 88.7  \scalebox{0.7}{ $\pm$ 2.8} & 87.4  \scalebox{0.7}{ $\pm$ 5.7} &   97.2 \scalebox{0.7}{ $\pm$ 1.3} & 95.5  \scalebox{0.7}{ $\pm$ 2.3} & -- & 95.8  \scalebox{0.7}{ $\pm$ 5.0} & 88.3  \scalebox{0.7}{ $\pm$ 4.5}&  98.0 \scalebox{0.7}{ $\pm$ 0.7} \\
      & grid & 99.9  \scalebox{0.7}{ $\pm$ 0.1} & -- & 100.0  \scalebox{0.7}{ $\pm$ 0.0} & 98.6  \scalebox{0.7}{ $\pm$ 0.8} & 100.0 \scalebox{0.7}{ $\pm$ 0.0}& 99.2  \scalebox{0.7}{ $\pm$ 0.1} &-- & 98.4  \scalebox{0.7}{ $\pm$ 0.7} & 92.5  \scalebox{0.7}{ $\pm$ 2.0}&  99.4 \scalebox{0.7}{ $\pm$ 0.0} \\
      & leather& 99.9  \scalebox{0.7}{ $\pm$ 0.1} & -- & 99.9  \scalebox{0.7}{ $\pm$ 0.1} & 100.0  \scalebox{0.7}{ $\pm$ 0.0} &  100.0 \scalebox{0.7}{ $\pm$ 0.0} & 99.5  \scalebox{0.7}{ $\pm$ 0.1} & -- & 99.3  \scalebox{0.7}{ $\pm$ 0.5} & 99.5  \scalebox{0.7}{ $\pm$ 0.1}&  99.7 \scalebox{0.7}{ $\pm$ 0.0} \\
      & tile & 100.0  \scalebox{0.7}{ $\pm$ 0.0} & -- & 100.0  \scalebox{0.7}{ $\pm$ 0.0} & 100.0  \scalebox{0.7}{ $\pm$ 0.0}&   99.9 \scalebox{0.7}{ $\pm$ 0.1}  & 99.3  \scalebox{0.7}{ $\pm$ 0.0} & -- & 98.5  \scalebox{0.7}{ $\pm$ 0.4} & 95.6  \scalebox{0.7}{ $\pm$ 2.1}&  98.2 \scalebox{0.7}{ $\pm$ 0.4} \\
      & wood & 97.5  \scalebox{0.7}{ $\pm$ 1.5} & -- & 91.4  \scalebox{0.7}{ $\pm$ 4.3} & 91.4  \scalebox{0.7}{ $\pm$ 2.5}&  98.0 \scalebox{0.7}{ $\pm$ 0.3} & 90.7  \scalebox{0.7}{ $\pm$ 1.9} & -- & 86.5  \scalebox{0.7}{ $\pm$ 3.9} & ?856  \scalebox{0.7}{ $\pm$ 2.0}&  92.4 \scalebox{0.7}{ $\pm$ 0.6} \\ \cmidrule(l){2-12}
      & average & 98.6  \scalebox{0.7}{ $\pm$ 0.3} & -- & 96.0  \scalebox{0.7}{ $\pm$ 0.7} & 95.5  \scalebox{0.7}{ $\pm$ 1.5}&  99.0 \scalebox{0.7}{ $\pm$ 0.3} & 96.8  \scalebox{0.7}{ $\pm$ 0.7} & -- & 95.7  \scalebox{0.7}{ $\pm$ 1.6} & 92.3  \scalebox{0.7}{ $\pm$ 1.1} &  97.5 \scalebox{0.7}{ $\pm$ 0.1} \\ \midrule
\multicolumn{2}{c}{overall average} & 97.2  \scalebox{0.7}{ $\pm$ 0.3} & -- & 95.3  \scalebox{0.7}{ $\pm$ 0.5} & 91.4  \scalebox{0.7}{ $\pm$ 0.5} &   97.0 \scalebox{0.7}{ $\pm$ 0.2} & 96.3  \scalebox{0.7}{ $\pm$ 0.4} & -- & 94.1  \scalebox{0.7}{ $\pm$ 0.8} & 92.3  \scalebox{0.7}{ $\pm$ 0.5}&  95.7 \scalebox{0.7}{ $\pm$ 0.2} \\
\bottomrule
\end{tabular}}
\end{table}

\subsection{Per-region overlap}
The $\text{AU-PRO}_{0.3}$ metric is defined as the area under the per-region overlap (PRO) curve for false positive rates up to 30 \% \cite{BergmannPaul2021TMAD}. To calculate PRO, we decompose the ground-truth label maps into $M$ connected components such that $C_{j,k}$ gives the set of anomalous pixels in a connected component $k$ of label map $j$. Let $P_{j}$ denote the predicted anomalous pixels when using a threshold $t$. \cite{BergmannPaul2021TMAD} defines PRO as:
\begin{equation}
    \text{PRO} = \frac{1}{M} \sum_j\sum_k \frac{\left\vert P_{j} \cap C_{j,k}\right\vert}{\left\vert C_{j,k}\right\vert}
\end{equation}
Unlike pixel-level AUROC, $\text{AU-PRO}$ assigns equal weight to small and large anomalies. This is desirable for practical applications where precise localization of small anomalies is at least as important as localization of large anomalies. \par In \cref{tab:mvtec_pro} we report $\text{AU-PRO}_{0.3}$ scores for our models from \cref{tab:mvtec_pixel_auroc} as well as the scores for PaDiM \cite{defard2020padim} for reference. Note that unlike our method, PaDiM relies on ImageNet pretraining, so this is not a fair comparison. The authors of CutPaste \cite{li2021cutpaste} and DRAEM \cite{zavrtanik2021draem} did not report $\text{AU-PRO}$ for their method but we hope that future methods that learn from scratch can compare their localization performance to our AU-PRO scores.\par 

\begin{table}[H]
\caption{$\text{AU-PRO}_{0.3}$ \% for MVTec AD defect localization and standard error across five different random seeds. Scores are calculated for $256\times 256$ resampled image and mask. Best scores between  PaDiM-WR50-Rd550 \cite{defard2020padim} and \placeholder{} within standard error are bold-faced. Note that PaDiM uses pretrained ImageNet features.}
    \label{tab:mvtec_pro}
    \centering
\resizebox{0.9\linewidth}{!}{%
\begin{tabular}{llccccccc}
\toprule
& & \multicolumn{1}{c}{SOTA} & \multicolumn{6}{c}{Our Experiments} \\ \cmidrule(r{2pt}){3-3}\cmidrule(l{2pt}){4-9}
       &      & \begin{tabular}{@{}c@{}} PaDiM \\ \cite{defard2020padim} \end{tabular}  &  \begin{tabular}{@{}c@{}} CutPaste \\ (end-to-end)  \end{tabular} &  FPI  &  PII & \begin{tabular}{@{}c@{}} \placeholder{} \\ (binary)  \end{tabular} &    \begin{tabular}{@{}c@{}} \placeholder{} \\ (continuous) \end{tabular} & \begin{tabular}{@{}c@{}} \placeholder{} \\ (logistic) \end{tabular}  \\
\midrule
\multirow{11}{*}{\rotatebox[origin=c]{90}{object}} & bottle &                          \textbf{94.8} &                   91.2 & 66.0 &         79.0 &   93.0  \scalebox{0.7}{ $\pm$ 0.9} &   89.9  \scalebox{0.7}{ $\pm$ 1.1} &     92.9  \scalebox{0.7}{ $\pm$ 0.3} \\
      & cable &                          88.8 &                   59.8 & 51.9 &         55.7 &   \textbf{87.6}  \scalebox{0.7}{ $\pm$ 3.4} &   85.4  \scalebox{0.7}{ $\pm$ 2.1} &     \textbf{89.9}  \scalebox{0.7}{ $\pm$ 1.0} \\
      & capsule &                          \textbf{93.5} &                   83.5 & 79.9 &         67.6 &  91.8  \scalebox{0.7}{ $\pm$ 0.8} &   79.9  \scalebox{0.7}{ $\pm$ 9.0} &     \textbf{91.4}  \scalebox{0.7}{ $\pm$ 2.2} \\
      & hazelnut &                        92.6 &                   81.3 & 71.4 &         90.9 &   \textbf{93.6}  \scalebox{0.7}{ $\pm$ 0.4} &   \textbf{93.1}  \scalebox{0.7}{ $\pm$ 1.3} &     \textbf{93.6}  \scalebox{0.7}{ $\pm$ 0.9} \\
      & metal nut &                          85.6 &                   54.4 & 72.2 &         91.5 &   \textbf{94.9}  \scalebox{0.7}{ $\pm$ 0.2} &   90.8  \scalebox{0.7}{ $\pm$ 1.1} &     \textbf{94.6}  \scalebox{0.7}{ $\pm$ 0.6} \\
      & pill &                          92.7 &                   83.1 & 50.4 &         65.2 &   93.7  \scalebox{0.7}{ $\pm$ 0.9} &   \textbf{92.5}  \scalebox{0.7}{ $\pm$ 3.5} &     \textbf{96.0}  \scalebox{0.7}{ $\pm$ 0.5} \\
      & screw &                          \textbf{94.4} &                   72.6 & 69.8 &         78.4 &   90.6  \scalebox{0.7}{ $\pm$ 1.3} &  80.6  \scalebox{0.7}{ $\pm$ 10.3} &     90.1  \scalebox{0.7}{ $\pm$ 0.3} \\
      & toothbrush &                          \textbf{93.1} &                   88.1 & 60.3 &         66.8 &   91.2  \scalebox{0.7}{ $\pm$ 0.6} &   89.0  \scalebox{0.7}{ $\pm$ 1.8} &     90.7  \scalebox{0.7}{ $\pm$ 1.0} \\
      & transistor &                          \textbf{84.5} &                   68.5 & 55.4 &         57.4 &   72.6  \scalebox{0.7}{ $\pm$ 4.4} &   63.3  \scalebox{0.7}{ $\pm$ 1.2} &     75.3  \scalebox{0.7}{ $\pm$ 2.4} \\
      & zipper &                          \textbf{95.9} &                   84.9 & 81.2 &         86.6 &   88.9  \scalebox{0.7}{ $\pm$ 0.5} &   83.6  \scalebox{0.7}{ $\pm$ 3.3} &     89.2  \scalebox{0.7}{ $\pm$ 0.3} \\ \cmidrule(l){2-9}
      & average &                          \textbf{91.6} &                   76.7 & 65.8 &         73.9 &   89.8  \scalebox{0.7}{ $\pm$ 0.8} &   84.8  \scalebox{0.7}{ $\pm$ 2.8} &     90.4  \scalebox{0.7}{ $\pm$ 0.5} \\ \midrule
\multirow{6}{*}{\rotatebox[origin=c]{90}{texture}} & carpet &                          \textbf{96.2} &                   50.4 & 21.6 &         93.5 &  84.0  \scalebox{0.7}{ $\pm$ 11.8} &   71.1  \scalebox{0.7}{ $\pm$ 8.2} &     85.0  \scalebox{0.7}{ $\pm$ 6.2} \\
      & grid &                          94.6 &                   91.5 & 86.0 &         95.9 &   \textbf{96.5}  \scalebox{0.7}{ $\pm$ 0.1} &   94.2  \scalebox{0.7}{ $\pm$ 0.8} &    \textbf{96.8}  \scalebox{0.7}{ $\pm$ 0.4} \\
      & leather &                          97.8 &                   83.7 & 84.1 &         98.1 &   \textbf{98.9}  \scalebox{0.7}{ $\pm$ 0.1} &   \textbf{98.6}  \scalebox{0.7}{ $\pm$ 0.4} &     \textbf{98.7}  \scalebox{0.7}{ $\pm$ 0.1} \\
      & tile &                          86.0 &                   54.4 & 42.0 &         83.2 &   \textbf{93.9}  \scalebox{0.7}{ $\pm$ 0.9} &   90.3  \scalebox{0.7}{ $\pm$ 2.5} &     \textbf{95.3}  \scalebox{0.7}{ $\pm$ 0.5} \\
      & wood &                          \textbf{91.1} &                   64.0 & 41.7 &         81.7 &   \textbf{89.2}  \scalebox{0.7}{ $\pm$ 2.4} &   \textbf{86.1}  \scalebox{0.7}{ $\pm$ 5.7} &     85.3  \scalebox{0.7}{ $\pm$ 3.7} \\ \cmidrule(l){2-9}
      & average &                          \textbf{93.2} &                   68.8 & 55.1 &         90.5 &   \textbf{92.5}  \scalebox{0.7}{ $\pm$ 2.0} &   88.1  \scalebox{0.7}{ $\pm$ 1.3} &     \textbf{92.2}  \scalebox{0.7}{ $\pm$ 1.4} \\ \midrule
\multicolumn{2}{c}{overall average} &                          \textbf{92.1} &                   74.1 & 62.3 &         79.4 &   90.7  \scalebox{0.7}{ $\pm$ 0.4} &   85.9  \scalebox{0.7}{ $\pm$ 2.1} &     91.0  \scalebox{0.7}{ $\pm$ 0.6} \\
\bottomrule
\end{tabular}}
\end{table}
\end{document}